\documentclass{article}
\usepackage[T1]{fontenc}
\usepackage{iclr2027_conference,times}

\usepackage{amsmath,amsfonts,bm}

\def\eqref#1{equation~\ref{#1}}

\def\1{\bm{1}}

\DeclareMathAlphabet{\mathsfit}{\encodingdefault}{\sfdefault}{m}{sl}
\SetMathAlphabet{\mathsfit}{bold}{\encodingdefault}{\sfdefault}{bx}{n}

\usepackage{hyperref,url,booktabs,multirow,array,xcolor,colortbl,tabularx,graphicx,adjustbox}
\usepackage{pifont,placeins,marvosym}
\newcommand{\xmark}{\textcolor{red!70!black}{\ding{55}}}
\newcommand{\cmark}{\textcolor{green!60!black}{\ding{51}}}
\hypersetup{colorlinks=false,pdfborder={0 0 1},citebordercolor={0 1 0}}
\definecolor{FamilyGray}{HTML}{F1F3F5}
\title{When History Helps and Hurts: Selective\\History Use across Multimodal Turns}
\newcommand{\AuthorOne}{Shuoyang Sun}
\newcommand{\AuthorTwo}{Kerui Gu}
\newcommand{\AuthorThree}{Hao Fang}
\newcommand{\AuthorFour}{Shaoli Huang}
\newcommand{\AuthorFive}{Bin Chen}
\newcommand{\InstitutionOne}{Harbin Institute of Technology, Shenzhen}
\newcommand{\InstitutionTwo}{AgiBot}
\newcommand{\InstitutionThree}{Tsinghua Shenzhen International Graduate School, Tsinghua University}

\makeatletter
\let\ReTurnOriginalFNSymbol\@fnsymbol
\renewcommand{\@fnsymbol}[1]{%
  \ifnum#1=3\relax\hbox{\Letter}\else\ReTurnOriginalFNSymbol{#1}\fi}
\makeatother

\author{%
\parbox[t]{\dimexpr\textwidth-2\tabcolsep\relax}{\centering
\AuthorOne\textsuperscript{1}\thanks{Equal contribution.}%
\hspace{0.8em}\AuthorTwo\textsuperscript{2}\footnotemark[1]%
\hspace{0.8em}\AuthorThree\textsuperscript{3}%
\hspace{0.8em}\AuthorFour\textsuperscript{2}\thanks{Project lead.}%
\hspace{0.8em}\AuthorFive\textsuperscript{1}\thanks{Corresponding author.}%
\\[0.8ex]
{\normalfont\mdseries
\textsuperscript{1}\InstitutionOne\hspace{1.5em}
\textsuperscript{2}\InstitutionTwo\\
\textsuperscript{3}\InstitutionThree}
}}

\iclrfinalcopy
\begin{document}
\maketitle
\lhead{Preprint}
\begin{abstract}
Reliable multimodal interaction depends on selective use of conversational history: an earlier question may remain relevant while its previous answer is outdated, whereas a current request may depend on historical evidence despite conflicting new observations. Existing multi-turn evaluations rarely separate these history-use demands from underlying question difficulty. To address this gap, we introduce ReTurn, a benchmark of 7,000 base tasks spanning visual and audio evidence for evaluating selective history use. For task-carrying history, Reconfirm/Reground require applying a historical question to current media while varying historical agreement; for evidence-carrying history, Retrieve/Rebind require answering a current question using historical media while varying current-media competition. Each pair preserves the target question, media, and answer. Tasks support open-ended and multiple-choice evaluation, with matched single-turn counterparts serving as answerability references. Across 13 omni-modal, vision-language, and audio-language models, median model-level open-ended accuracy falls from 93.7\% with direct input to 72.3\% in conversation. Behavioral probes show that high question recall can coexist with weaker task application, while competing media can redirect answers away from historical targets. Supervised adaptation yields only partial gains. ReTurn provides a controlled framework for assessing whether multimodal models select and use the historical information required by each request. Project page: \mbox{\url{https://link-world.github.io/ReTurn/}}.
\end{abstract}


\section{Introduction}
\label{sec:introduction}

Multimodal large language models (MLLMs) increasingly support sustained interaction across visual and audio inputs. Vision-language, audio-language, and omni-modal systems allow users to discuss successive observations, revisit earlier media, and continue a task without restating it~\citep{DBLP:journals/corr/abs-2511-21631,DBLP:conf/nips/GhoshGKKKLYDMVC25,DBLP:journals/corr/abs-2509-17765,tong2025interactiveomni}. Evaluation has accordingly expanded to multi-image dialogue and streaming audiovisual interaction~\citep{DBLP:conf/nips/LiuCZWDZLX0L024,DBLP:conf/cvpr/WangWCW0Z25}. In these settings, conversational history supplies questions and evidence that may be absent from the current turn but remain essential to answering it.

\begin{figure*}[t]
\centering
\includegraphics[width=\textwidth]{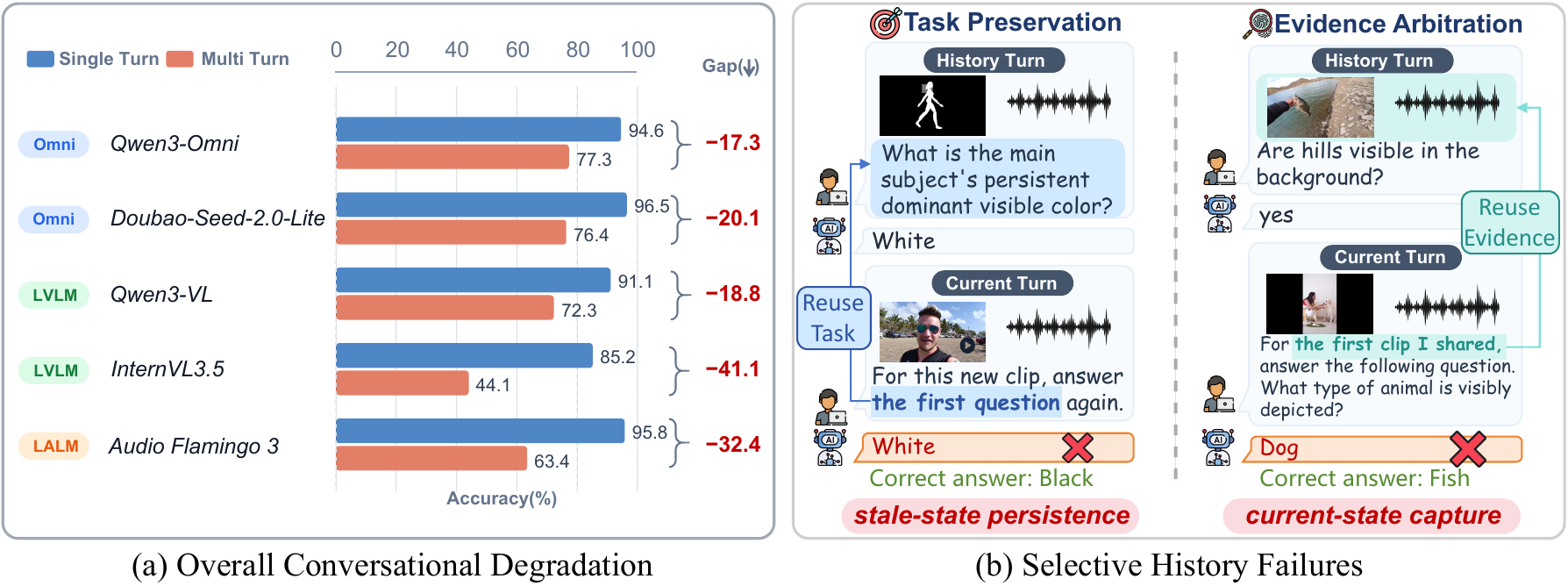}
\caption{\textbf{Selective history use in ReTurn.} (a) Single-turn versus multi-turn OpenQA accuracy across model families, evaluated on supported modalities. (b) Illustrative failures in applying a historical question to current media (Reground) and answering a current question from historical media despite competition (Rebind).}
\label{fig:overview}
\end{figure*}

However, history is not uniformly beneficial. When a user applies an earlier question to new media, the question remains necessary, but its previous answer may be outdated. Conversely, a request about earlier media requires using historical evidence despite conflicting current observations. These complementary demands call for \textbf{selective history use} (Figure~\ref{fig:overview}). Existing evaluations examine conversational memory, information updating, and multi-turn performance~\citep{DBLP:conf/iclr/WuWYZCY25,DBLP:conf/acl/MaharanaLTBBF24,DBLP:conf/acl/XueTHLHLLXZF0RG25,laban2026llms}. Yet assessing this ability requires distinguishing history-use difficulties from underlying question difficulty, while testing how misleading context affects the use of necessary history.

To address this need, we introduce \textbf{ReTurn}, a benchmark for evaluating selective history use through two roles of history. \emph{Task-carrying history} supplies a question to apply to current media; \emph{Reconfirm/Reground} vary whether historical media supports the same answer or a different one. \emph{Evidence-carrying history} supplies the media needed to answer the current question; \emph{Retrieve/Rebind} vary competition from current non-target media. Each local pair preserves the target question, media, and correct answer, while a matched single-turn counterpart assesses answerability with the question and target media supplied directly. ReTurn comprises 7,000 base tasks spanning visual and audio evidence, with paired open-ended (OpenQA) and multiple-choice (MCQ) views. Evaluation retains each model's historical replies, allowing earlier errors to remain part of the context.

Models often answer the underlying questions reliably with direct input yet struggle in conversation. Across 13 omni-modal models, large vision-language models (LVLMs), and large audio-language models (LALMs), median model-level OpenQA accuracy falls from 93.7\% in Single-turn to 72.3\% in Multi-turn. Long conversations expose complementary difficulties: applying a historical question despite its outdated answer (Reground versus Reconfirm), and preserving a historical target despite current competitors (Rebind versus Retrieve).

These findings motivate two hypothesized demands: \emph{history access}, recovering required historical information, and \emph{cross-turn binding}, pairing questions with target evidence across turns despite competing within-turn associations. Behavioral probes show high question recall alongside weaker application, while swapping current competitors can redirect answers away from unchanged historical targets. Inference-time assistance and supervised adaptation yield only partial gains, motivating evaluation of both historical information recovery and its use under interference.

Our contributions are threefold:
\begin{itemize}
\setlength{\itemsep}{0pt}
\item[$\bullet$] We identify selective history use as a requirement for multimodal interaction, highlighting the dual role of history as essential information and a potential source of interference.
\item[$\bullet$] We introduce ReTurn, an omni-modal benchmark that distinguishes task-carrying from evidence-carrying history, with matched pairs and single-turn references.
\item[$\bullet$] We evaluate omni-modal models, LVLMs, and LALMs, revealing limitations in selective history use and investigating failures through behavioral diagnostics and mitigation studies.
\end{itemize}

\section{Related Work}
\label{sec:related}

\subsection{Multimodal Models for Interactive Systems}

Multimodal models have broadened the range of evidence available to conversational assistants. Vision-language models such as Qwen3-VL and InternVL3.5 support image and video understanding~\citep{DBLP:journals/corr/abs-2511-21631,DBLP:journals/corr/abs-2508-18265}, while Audio Flamingo~3 and MiMo-Audio support speech and general audio~\citep{DBLP:conf/nips/GhoshGKKKLYDMVC25,DBLP:journals/corr/abs-2512-23808}. Alongside these modality-specific systems, omni-modal models integrate visual, acoustic, and textual processing, while InteractiveOmni explicitly targets audiovisual multi-turn dialogue~\citep{DBLP:journals/corr/abs-2509-17765,DBLP:journals/corr/abs-2604-27393,tong2025interactiveomni}. These interfaces allow successive requests to build on previously supplied media rather than isolated inputs. As interaction extends across turns, these capabilities also require models to reuse earlier questions and observations according to the user's current request.

\subsection{Multi-turn Evaluation and History Use}

ConvBench, MMDU, and MultiVerse evaluate multi-turn visual dialogue~\citep{DBLP:conf/nips/LiuYZYLZL00SZ24,DBLP:conf/nips/LiuCZWDZLX0L024,DBLP:conf/iccv/LeeLZHKKYRJHKC25}, while MT-Video-Bench, Audio MultiChallenge, and OmniMMI cover video, audio, and streaming audiovisual interaction~\citep{DBLP:conf/acl/PanXZWWZHPHGLPL26,DBLP:conf/acl/GosaiVTLYBUFJLH26,DBLP:conf/cvpr/WangWCW0Z25}. LoCoMo, LongMemEval, and MMRC study conversational memory~\citep{DBLP:conf/acl/MaharanaLTBBF24,DBLP:conf/iclr/WuWYZCY25,DBLP:conf/acl/XueTHLHLLXZF0RG25}. \citet{laban2026llms} compare single- and multi-turn performance with matched task content. ReTurn examines history as both necessary information and potential interference: matched pairs preserve the dependency and target while varying historical agreement or current-media competition. Single-turn references assess target answerability (Table~\ref{tab:benchmark_comparison}).

\begin{table*}[t]
\centering
\small
\setlength{\tabcolsep}{5.2pt}
\renewcommand{\arraystretch}{1.1}

\caption{Representative multi-turn benchmarks. Scale uses native units (q.: questions; dlg.: dialogues); counts are not directly comparable. Checks denote explicit designs: \textbf{Update}, revised information; \textbf{Direct}, matched direct-input reference; \textbf{Role}, separate task-/evidence-carrying history; \textbf{Conflict}, target-preserving reference/conflict pairs.}
\label{tab:benchmark_comparison}

\begin{tabularx}{\textwidth}{
>{\raggedright\arraybackslash}p{0.22\textwidth}
>{\centering\arraybackslash}X
>{\centering\arraybackslash}p{0.14\textwidth}
*{4}{>{\centering\arraybackslash}X}
}

\toprule

\multirow{2}{*}{\textbf{Benchmark}}
& \multicolumn{2}{c}{\textbf{Scope}}
& \multicolumn{2}{c}{\textbf{Structure}}
& \multicolumn{2}{c}{\textbf{History Design}}
\\

\cmidrule(lr){2-3}
\cmidrule(lr){4-5}
\cmidrule(lr){6-7}

& \textbf{Modality}
& \textbf{Scale}
& \textbf{Update}
& \textbf{Direct}
& \textbf{Role}
& \textbf{Conflict}
\\

\midrule

LongMemEval
& T
& 500 q.
& \cmark
& \xmark
& \xmark
& \xmark
\\

Lost in Conversation
& T
& 600 tasks
& \xmark
& \cmark
& \xmark
& \xmark
\\

\midrule

MMDU
& V+T
& 110 dlg.
& \xmark
& \xmark
& \xmark
& \xmark
\\

MultiVerse
& V+T
& 647 dlg.
& \xmark
& \xmark
& \xmark
& \xmark
\\

MMRC
& V+T
& 5,120 dlg.
& \cmark
& \xmark
& \xmark
& \xmark
\\

MT-Video-Bench
& V+T
& 1,000 dlg.
& \xmark
& \xmark
& \xmark
& \xmark
\\

Audio MultiChallenge
& A+T
& 452 dlg.
& \cmark
& \xmark
& \xmark
& \xmark
\\

OmniMMI
& V+A+T
& 2,290 q.
& \xmark
& \xmark
& \xmark
& \xmark
\\

\midrule

\rowcolor{gray!10}
\textbf{ReTurn (Ours)}
& \textbf{V+A+T}
& \mbox{\textbf{7,000 tasks}}
& \cmark
& \cmark
& \cmark
& \cmark
\\

\bottomrule

\end{tabularx}
\end{table*}

\section{ReTurn: Benchmarking Selective History Use}
\label{sec:benchmark}

\begin{figure*}[t]
\centering
\includegraphics[width=\textwidth]{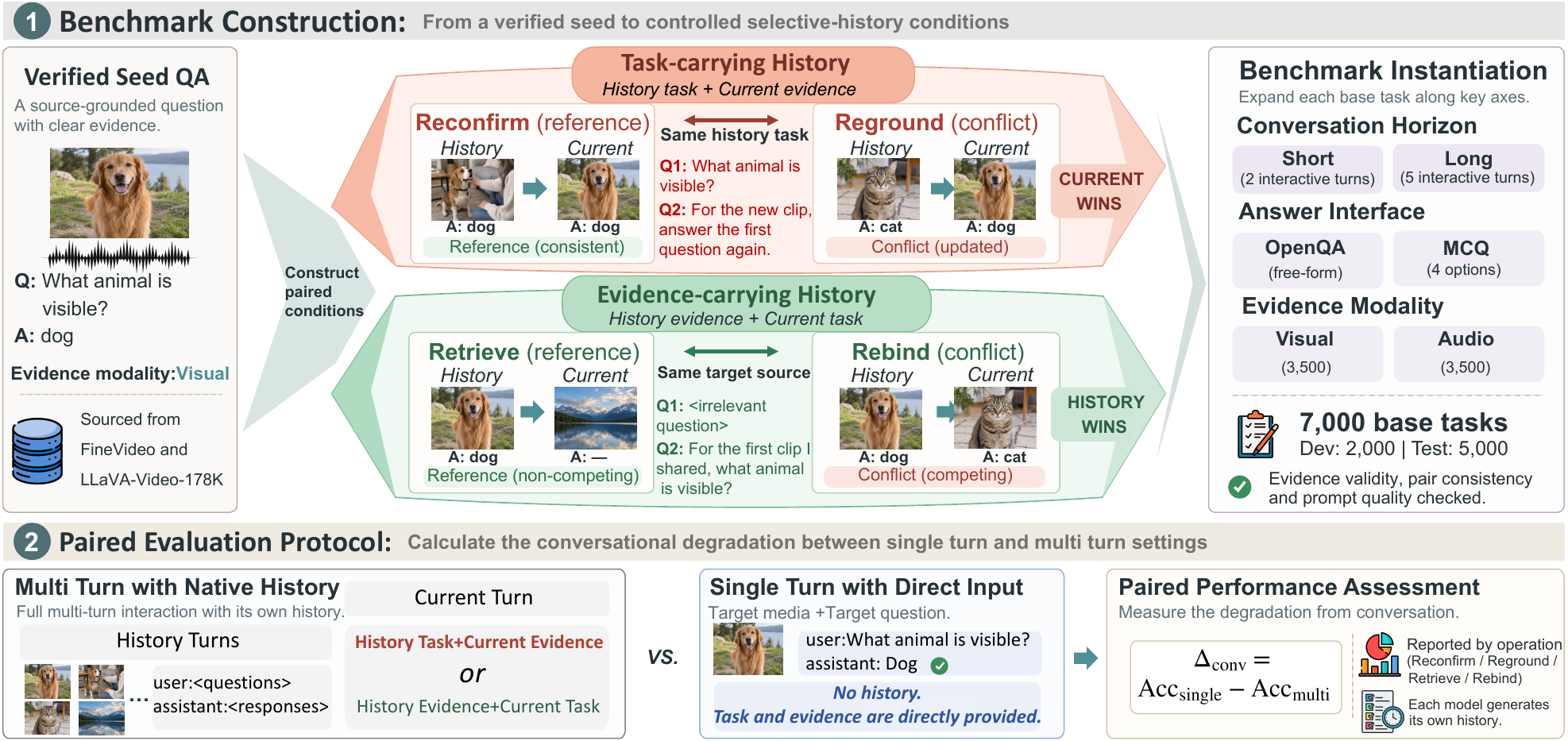}
\caption{\textbf{ReTurn construction and evaluation.} Verified media--question pairs yield matched task-carrying and evidence-carrying conditions. Media annotations indicate supported answers; evaluation uses model-generated histories and matched single-turn references.}
\label{fig:benchmark}
\end{figure*}

ReTurn evaluates selective history use through matched conversations built from verified media--question pairs. Each construction preserves the information needed to answer the final request while varying contextual agreement or competition (Figure~\ref{fig:benchmark}).

\subsection{Task Design}
\label{sec:history-roles}
\label{sec:matched-conditions}

Let $q_h,q_c$ denote historical/current questions and $m_h,m_c$ historical/current media. With $a(q,m)$ denoting the supported answer, the two required targets are:
\begin{equation}
y^\star_{\mathrm{task}}=a(q_h,m_c),\qquad y^\star_{\mathrm{evidence}}=a(q_c,m_h).
\label{eq:history-roles}
\end{equation}
\textbf{Task-carrying history.} Reconfirm and Reground ask the model to answer the first question about current target media. The final request and current target remain fixed, while the initial media support the target answer in Reconfirm and a different answer in Reground. The pair tests whether models preserve the question while updating its answer: Reconfirm supports answer reuse, whereas Reground requires resolving the new evidence.

\textbf{Evidence-carrying history.} Retrieve and Rebind provide a current question about historical target media alongside current non-target media. They preserve the target and preceding dialogue while replacing a weaker competitor (Retrieve) with media supporting a plausible different answer (Rebind). Both target the historical source, with Retrieve providing a lower-competition reference for Rebind.

Each pair fixes the target question and media, answer, options, historical dependency, and horizon. Pair construction uses media-supported answers to define agreement and competition, while evaluation retains native model replies. These matched contrasts assess selective history use (Appendix~\ref{app:provenance}).

\subsection{Dataset Construction}
\label{sec:construction}

\textbf{Construction and verification.} We localize clips from FineVideo~\citep{farre2024finevideo} and LLaVA-Video-178K~\citep{zhang2024videoinstructiontuningsynthetic}; source annotations guide retrieval rather than supply answer evidence. Doubao-Seed-2.0-Lite~\citep{seed2026seed2} verifies questions and answers against actual visual or acoustic content. We pair media by their answer relations; GPT-5.6 Sol~\citep{openai2026gpt56systemcard} assists clarity, consistency, and leakage checks. Questions target source-grounded facts; evidence-carrying histories use questions distinct from the final request to reduce answer copying. Provenance and automated audits appear in Appendices~\ref{app:provenance} and~\ref{app:scoring}.

\textbf{Coverage and task views.} ReTurn contains 7,000 tasks in 3,500 matched pairs, with 1,750 tasks per operation and balanced visual/audio evidence and Short/Long horizons. Evidence modality identifies the required channel, not input restriction. Short/Long contain two/five media-bearing user turns, including the final turn, and are not universally item-matched. OpenQA and four-option MCQ provide two views per task (14,000 total). Task-carrying MCQ inherits the original options and letter mapping.

\textbf{Development and partitioning.} Development, primarily with Qwen3-Omni, prioritized directly answerable questions and measurable history effects. These goals guided refinement without requiring universal correctness or large degradation for every task. The split comprises Dev (2,000: Dev-Train 1,500; Dev-Val 500) and Test (5,000), using pair and fact-group metadata rather than final model scores. Appendix~\ref{app:provenance} documents shared media and development exposure, which bound conclusions about unseen-source generalization.

\subsection{Evaluation Protocol}
\label{sec:return-protocol}

\textbf{Single-turn and multi-turn inputs.} Single-turn supplies the complete question and target media directly, including historical target media when needed. Multi-turn evaluates the final request after model-generated historical replies, separately for each interface. Earlier errors remain; changed historical media can also change assistant replies. Models consume supported modalities (Appendix~\ref{app:models}).

\textbf{Scoring and comparisons.} We score final-answer content accuracy for OpenQA and MCQ. Frozen rules provide initial scores; rejected responses receive GPT-5.6 Sol review blinded to model and condition. Uncertain judgments count as incorrect. Scoring rules and alternative-judge audits appear in Appendix~\ref{app:scoring}. The \emph{conversational degradation}, $\Delta_{\mathrm{conv}}=\mathrm{Acc}_{\mathrm{single}}-\mathrm{Acc}_{\mathrm{multi}}$, is reported in percentage points (pp). Within-pair comparisons assess historical agreement or current-media competition. Scores use valid single-turn/multi-turn response pairs; aggregation is detailed in Appendix~\ref{app:full-results}.

\section{Evaluating Selective History Use}
\label{sec:evaluation}

We examine whether strong single-turn performance translates into selective history use across omni-modal models, LVLMs, and LALMs. Table~\ref{tab:return-main-horizon} reports Test results for all 13 models on supported modalities (configurations: Appendix~\ref{app:models}); Figure~\ref{fig:profiles} summarizes paired conditions, evidence modality, and answer interface.
\begin{table*}[t]
\centering
\definecolor{RetWarm}{HTML}{B94F45}
\definecolor{RetCool}{HTML}{6689A3}
\definecolor{RetMuted}{HTML}{747A80}
\definecolor{RetText}{HTML}{2E3338}
\definecolor{OmniBand}{HTML}{F0F4F7}
\definecolor{OmniAccent}{HTML}{46667E}
\definecolor{VLMBand}{HTML}{EFF5F3}
\definecolor{VLMAccent}{HTML}{4C7368}
\definecolor{ALMBand}{HTML}{F7F3ED}
\definecolor{ALMAccent}{HTML}{786148}

\providecommand{\modelname}[1]{#1}
\renewcommand{\modelname}[1]{{\fontsize{7.6pt}{8.4pt}\selectfont #1}}
\providecommand{\dropc}[3]{#1}
\renewcommand{\dropc}[3]{\cellcolor{RetWarm!\number\numexpr#3*3/2\relax}#1{\scriptsize\,(\(\downarrow\)#2)}}
\providecommand{\risec}[3]{#1}
\renewcommand{\risec}[3]{\cellcolor{RetCool!#3}#1{\scriptsize\,(\(\uparrow\)#2)}}
\providecommand{\resultheader}{}
\renewcommand{\resultheader}{\multirow[c]{2}{=}{\centering\raisebox{-0.5ex}{{\normalsize\bfseries Model Name}}} & \multicolumn{5}{c}{\bfseries OpenQA} & \multicolumn{5}{c}{\bfseries MCQ} \\[-0.18ex]
\cmidrule(lr){2-6}\cmidrule(lr){7-11}
& \bfseries Overall & \bfseries Reconfirm & \bfseries Reground & \bfseries Retrieve & \bfseries Rebind & \bfseries Overall & \bfseries Reconfirm & \bfseries Reground & \bfseries Retrieve & \bfseries Rebind \\[0.02ex]}
\providecommand{\horizonpanel}[2]{}
\renewcommand{\horizonpanel}[2]{\multicolumn{11}{@{}l}{\hspace{0.08em}\textcolor{RetText}{\textbf{\textit{#1}}\,\ensuremath{\cdot}\,#2}}\\}
\providecommand{\familyband}[4]{}
\renewcommand{\familyband}[4]{\rowcolor{#1}\multicolumn{11}{@{}l}{\hspace{0.10em}{\scriptsize\textcolor{#2}{\bfseries #3}\,\ensuremath{\cdot}\,\textcolor{RetMuted}{#4}}}\\}

\caption{\textbf{ReTurn Test results across 13 models.} Cells report Multi-turn accuracy (\%) and the drop ($\downarrow$) or gain ($\uparrow$) from Single-turn in pp. Darker red indicates larger drops; blue indicates gains. Models are evaluated on supported modalities. Full Single-turn scores and valid-pair denominators appear in Appendix~\ref{app:full-results}.}
\label{tab:return-main-horizon}
\providecommand{\RetPanelTitle}[1]{{\bfseries\itshape #1}}
\small
\setlength{\tabcolsep}{3.10pt}
\renewcommand{\arraystretch}{1.05}
\begin{adjustbox}{max width=\textwidth}
\begin{tabular}{@{}>{\raggedright\arraybackslash}m{3.10cm}@{\hspace{0.58em}}ccc@{\hspace{0.30em}}cc@{\hspace{0.88em}}ccc@{\hspace{0.30em}}cc@{}}
\toprule
\resultheader
\midrule
\multicolumn{11}{@{}l}{\hspace{0.08em}\textcolor{RetText}{\RetPanelTitle{(a) Short conversations}\,\ensuremath{\cdot}\,2 native interactive turns}}\\
\midrule
\familyband{OmniBand}{OmniAccent}{OMNI}{Visual + audio evidence}
\modelname{Qwen3.8-Omni-Flash} & \dropc{91.1}{3.4}{3} & \dropc{94.4}{1.8}{3} & \dropc{95.0}{1.1}{2} & \risec{94.4}{1.4}{2} & \dropc{80.6}{12.3}{6} & \dropc{93.4}{5.4}{4} & \dropc{98.6}{0.0}{2} & \dropc{93.9}{4.6}{4} & \dropc{99.0}{0.0}{2} & \dropc{81.9}{17.1}{8}\\
\modelname{Qwen3-Omni} & \dropc{91.6}{3.0}{3} & \risec{95.4}{0.3}{2} & \dropc{91.0}{4.0}{3} & \dropc{94.2}{0.0}{2} & \dropc{85.9}{8.3}{5} & \dropc{89.7}{9.4}{5} & \dropc{97.0}{1.6}{3} & \dropc{67.4}{31.2}{12} & \dropc{99.2}{0.3}{2} & \dropc{95.2}{4.3}{3}\\
\modelname{Doubao-Seed-2.0-Lite} & \dropc{89.8}{6.6}{4} & \dropc{95.4}{2.1}{3} & \dropc{93.1}{4.3}{3} & \dropc{94.7}{0.6}{2} & \dropc{75.8}{19.5}{9} & \dropc{89.0}{10.4}{5} & \dropc{98.7}{0.6}{2} & \dropc{71.8}{27.5}{11} & \dropc{98.9}{0.6}{2} & \dropc{86.7}{12.8}{6}\\
\modelname{MiniCPM-o 4.5} & \dropc{78.7}{14.4}{7} & \dropc{88.5}{5.4}{4} & \dropc{80.8}{13.1}{6} & \dropc{91.0}{1.3}{2} & \dropc{54.4}{37.9}{15} & \dropc{68.4}{29.2}{12} & \dropc{96.0}{1.3}{2} & \dropc{32.3}{65.0}{24} & \dropc{96.5}{1.4}{2} & \dropc{49.0}{49.0}{18}\\
\modelname{MiMo-V2.5} & \dropc{80.2}{11.2}{6} & \dropc{90.1}{2.6}{3} & \dropc{79.5}{13.1}{6} & \dropc{87.7}{2.6}{3} & \dropc{63.5}{26.7}{11} & \dropc{75.9}{21.7}{9} & \dropc{89.0}{8.8}{5} & \dropc{46.9}{50.9}{19} & \dropc{96.3}{1.1}{2} & \dropc{71.4}{26.1}{11}\\
\modelname{Nemotron-3-Nano-Omni} & \dropc{85.9}{5.2}{4} & \risec{92.5}{2.4}{3} & \dropc{75.4}{14.7}{7} & \risec{93.4}{1.4}{2} & \dropc{82.2}{9.8}{5} & \dropc{78.7}{19.0}{8} & \risec{99.0}{1.1}{2} & \dropc{34.4}{63.5}{23} & \risec{99.4}{1.9}{3} & \dropc{82.1}{15.4}{7}\\[0.8ex]
\familyband{VLMBand}{VLMAccent}{LVLM}{Visual evidence}
\modelname{Qwen3.7-Plus} & \dropc{92.6}{0.9}{2} & \dropc{94.6}{0.3}{2} & \dropc{92.0}{2.9}{3} & \risec{93.3}{1.3}{2} & \dropc{90.4}{1.6}{3} & \dropc{93.5}{4.7}{4} & \dropc{98.4}{0.3}{2} & \dropc{80.1}{18.6}{8} & \risec{98.1}{0.3}{2} & \dropc{97.4}{0.3}{2}\\
\modelname{Qwen3-VL} & \dropc{87.9}{3.1}{3} & \dropc{90.7}{1.3}{2} & \dropc{87.8}{4.2}{3} & \risec{90.7}{0.6}{2} & \dropc{82.4}{7.7}{5} & \dropc{80.3}{17.6}{8} & \risec{98.7}{1.6}{3} & \dropc{32.1}{65.1}{24} & \dropc{98.4}{0.3}{2} & \dropc{92.0}{6.7}{4}\\
\modelname{MiniCPM-V 4.5} & \dropc{82.3}{6.8}{4} & \risec{88.8}{1.3}{2} & \dropc{60.9}{26.6}{11} & \risec{91.7}{1.0}{2} & \dropc{87.9}{2.9}{3} & \dropc{76.4}{20.7}{9} & \dropc{94.2}{2.2}{3} & \dropc{20.8}{75.6}{27} & \dropc{97.8}{0.0}{2} & \dropc{92.7}{5.1}{4}\\
\modelname{InternVL3.5} & \dropc{56.4}{28.7}{12} & \dropc{42.9}{37.8}{15} & \dropc{29.8}{51.0}{19} & \dropc{87.5}{1.9}{3} & \dropc{65.1}{24.4}{10} & \dropc{70.5}{24.6}{10} & \dropc{93.3}{1.0}{2} & \dropc{11.9}{82.4}{29} & \risec{97.8}{1.9}{3} & \dropc{78.9}{16.9}{8}\\[0.8ex]
\familyband{ALMBand}{ALMAccent}{LALM}{Audio evidence}
\modelname{StepAudio 3} & \dropc{80.0}{18.7}{8} & \risec{99.0}{0.6}{2} & \risec{99.4}{1.0}{2} & \dropc{87.5}{11.6}{6} & \dropc{34.1}{65.0}{24} & \dropc{84.8}{15.2}{7} & \dropc{100.0}{0.0}{2} & \dropc{100.0}{0.0}{2} & \dropc{99.7}{0.3}{2} & \dropc{39.5}{60.5}{22}\\
\modelname{Audio Flamingo 3} & \dropc{72.6}{23.2}{10} & \dropc{95.5}{2.2}{3} & \dropc{47.3}{50.5}{19} & \dropc{93.3}{0.6}{2} & \dropc{54.5}{39.4}{15} & \dropc{65.8}{34.0}{13} & \risec{100.0}{0.3}{2} & \dropc{6.4}{93.3}{33} & \dropc{99.4}{0.6}{2} & \dropc{57.7}{42.3}{16}\\
\modelname{MiMo-Audio} & \dropc{62.7}{32.2}{13} & \dropc{86.3}{9.9}{5} & \dropc{72.5}{23.6}{10} & \dropc{73.1}{20.5}{9} & \dropc{18.9}{74.7}{27} & \dropc{49.6}{48.2}{18} & \dropc{97.4}{1.3}{2} & \dropc{13.1}{85.6}{31} & \dropc{66.0}{30.8}{12} & \dropc{21.8}{75.0}{27}\\
\midrule
\multicolumn{11}{@{}l}{\hspace{0.08em}\textcolor{RetText}{\RetPanelTitle{(b) Long conversations}\,\ensuremath{\cdot}\,5 native interactive turns}}\\
\midrule
\familyband{OmniBand}{OmniAccent}{OMNI}{Visual + audio evidence}
\modelname{Qwen3.8-Omni-Flash} & \dropc{83.2}{11.8}{6} & \dropc{91.3}{5.6}{4} & \dropc{73.1}{23.9}{10} & \risec{93.8}{0.8}{2} & \dropc{74.6}{18.4}{8} & \dropc{90.2}{8.6}{5} & \dropc{95.3}{3.2}{3} & \dropc{89.9}{8.7}{5} & \dropc{98.9}{0.2}{2} & \dropc{76.8}{22.2}{9}\\
\modelname{Qwen3-Omni} & \dropc{62.9}{31.7}{13} & \dropc{84.8}{10.1}{5} & \dropc{72.5}{22.4}{9} & \dropc{74.7}{19.5}{9} & \dropc{19.5}{74.7}{27} & \dropc{64.9}{34.3}{13} & \dropc{95.8}{3.0}{3} & \dropc{48.2}{50.7}{19} & \dropc{91.5}{8.0}{5} & \dropc{24.0}{75.5}{27}\\
\modelname{Doubao-Seed-2.0-Lite} & \dropc{63.0}{33.6}{13} & \dropc{82.9}{14.9}{7} & \dropc{77.9}{19.8}{9} & \dropc{76.5}{18.9}{8} & \dropc{14.7}{80.6}{29} & \dropc{66.3}{33.1}{13} & \dropc{82.2}{17.0}{8} & \dropc{62.6}{36.6}{14} & \dropc{91.2}{8.3}{5} & \dropc{29.1}{70.4}{25}\\
\modelname{MiniCPM-o 4.5} & \dropc{42.6}{50.5}{19} & \dropc{40.0}{53.8}{20} & \dropc{35.8}{57.9}{21} & \dropc{78.9}{13.4}{6} & \dropc{15.5}{76.8}{28} & \dropc{47.8}{49.8}{19} & \dropc{64.6}{32.6}{13} & \dropc{23.7}{73.6}{27} & \dropc{89.0}{9.0}{5} & \dropc{13.8}{84.2}{30}\\
\modelname{MiMo-V2.5} & \dropc{47.3}{44.7}{17} & \dropc{59.7}{34.1}{13} & \dropc{48.5}{45.3}{17} & \dropc{65.6}{24.6}{10} & \dropc{15.4}{74.9}{27} & \dropc{53.2}{44.2}{17} & \dropc{71.7}{25.6}{11} & \dropc{31.8}{65.4}{24} & \dropc{84.8}{12.6}{6} & \dropc{24.3}{73.1}{26}\\
\modelname{Nemotron-3-Nano-Omni} & \dropc{40.5}{50.8}{19} & \dropc{34.7}{55.8}{21} & \dropc{25.8}{64.8}{24} & \dropc{81.8}{10.2}{5} & \dropc{19.7}{72.3}{26} & \dropc{58.6}{39.0}{15} & \dropc{97.9}{0.0}{2} & \dropc{17.9}{80.0}{29} & \dropc{93.8}{3.7}{3} & \dropc{25.0}{72.5}{26}\\[0.8ex]
\familyband{VLMBand}{VLMAccent}{LVLM}{Visual evidence}
\modelname{Qwen3.7-Plus} & \dropc{81.4}{12.6}{6} & \dropc{89.4}{6.4}{4} & \dropc{64.1}{31.7}{13} & \risec{92.7}{0.6}{2} & \dropc{79.2}{12.8}{6} & \dropc{83.7}{14.4}{7} & \dropc{95.2}{3.2}{3} & \dropc{54.8}{43.6}{17} & \dropc{97.4}{0.3}{2} & \dropc{87.2}{10.5}{6}\\
\modelname{Qwen3-VL} & \dropc{56.7}{34.5}{13} & \dropc{80.1}{12.2}{6} & \dropc{63.8}{28.5}{12} & \dropc{60.1}{30.0}{12} & \dropc{23.0}{67.1}{24} & \dropc{59.5}{38.1}{15} & \dropc{93.9}{2.6}{3} & \dropc{32.1}{64.4}{23} & \dropc{84.3}{14.4}{7} & \dropc{27.8}{70.9}{26}\\
\modelname{MiniCPM-V 4.5} & \dropc{64.8}{24.3}{10} & \dropc{79.5}{8.0}{5} & \dropc{28.8}{58.7}{22} & \dropc{82.4}{8.3}{5} & \dropc{68.4}{22.4}{9} & \dropc{64.6}{32.1}{13} & \dropc{93.3}{2.2}{3} & \dropc{8.0}{87.5}{31} & \dropc{91.1}{6.7}{4} & \dropc{65.8}{31.9}{13}\\
\modelname{InternVL3.5} & \dropc{31.8}{53.4}{20} & \dropc{8.7}{72.4}{26} & \dropc{7.7}{73.4}{26} & \dropc{73.8}{15.7}{7} & \dropc{37.1}{52.4}{19} & \dropc{58.3}{36.9}{14} & \dropc{92.9}{1.6}{3} & \dropc{12.2}{82.4}{29} & \dropc{88.5}{7.3}{4} & \dropc{39.6}{56.2}{21}\\[0.8ex]
\familyband{ALMBand}{ALMAccent}{LALM}{Audio evidence}
\modelname{StepAudio 3} & \dropc{85.2}{13.7}{7} & \dropc{98.7}{0.0}{2} & \risec{99.0}{0.3}{2} & \dropc{91.0}{8.0}{5} & \dropc{51.9}{47.1}{18} & \dropc{81.4}{18.6}{8} & \dropc{100.0}{0.0}{2} & \dropc{100.0}{0.0}{2} & \dropc{97.1}{2.9}{3} & \dropc{28.5}{71.5}{26}\\
\modelname{Audio Flamingo 3} & \dropc{54.2}{41.7}{16} & \dropc{64.2}{33.5}{13} & \dropc{45.7}{52.1}{19} & \dropc{86.9}{7.1}{4} & \dropc{19.9}{74.0}{27} & \dropc{55.3}{44.6}{17} & \dropc{98.7}{1.0}{2} & \dropc{4.8}{94.9}{34} & \dropc{94.2}{5.8}{4} & \dropc{23.4}{76.6}{28}\\
\modelname{MiMo-Audio} & \dropc{37.4}{58.2}{21} & \dropc{50.8}{46.6}{18} & \dropc{42.2}{55.3}{20} & \dropc{52.2}{41.3}{16} & \dropc{4.2}{89.4}{32} & \dropc{34.6}{63.6}{23} & \dropc{58.8}{40.9}{16} & \dropc{18.8}{80.8}{29} & \dropc{53.8}{42.9}{16} & \dropc{7.1}{89.7}{32}\\
\bottomrule
\end{tabular}
\end{adjustbox}
\end{table*}

\subsection{Overall Results and the Access--Binding Hypothesis}
\label{sec:eval-overall}

Models reliably answer directly supplied questions and target media, yet struggle to resolve the same targets through history. Median model-level OpenQA accuracy falls from 93.7\% in Single-turn to 72.3\% in Multi-turn, with median $\Delta_{\mathrm{conv}}=20.1$ pp. This contrast exposes history-use difficulties beyond direct answerability.

Extended histories expose larger gaps. Overall Multi-turn accuracy is lower in Long than Short for 12 of 13 models on OpenQA and all 13 on MCQ. Median model-level $\Delta_{\mathrm{conv}}$ rises from 6.8 to 34.5 pp on OpenQA and from 19.0 to 36.9 pp on MCQ. Short and Long differ in both history composition and extent, so this is a configuration comparison rather than an isolated length effect.

\begin{samepage}
Conflict conditions expose weaknesses that reference conditions can conceal. In Long OpenQA, median accuracy falls from 79.5\% in Reconfirm to 48.5\% in Reground, and from 78.9\% in Retrieve to 19.9\% in Rebind. On both sides, success with relevant history does not ensure robustness to conflicting context.
\end{samepage}

\begin{figure}[t]
\centering
\includegraphics[width=.92\textwidth]{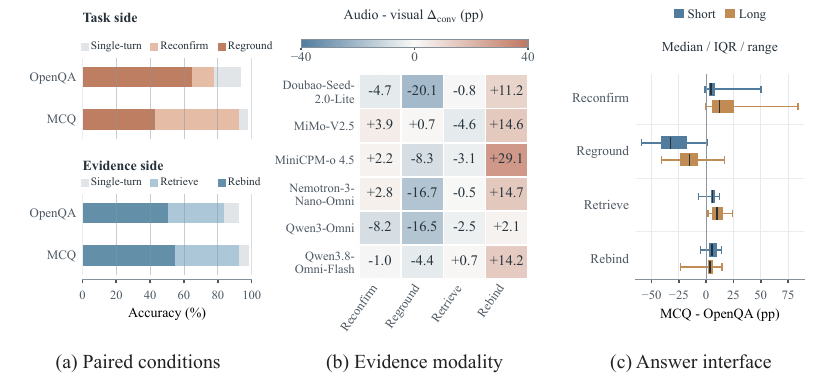}
\caption{\textbf{Paired conditions, modality, and answer interface.} (a) Mean accuracy on complete operation pairs across 13 models, pooling horizons; bars overlay accuracies. (b) Audio minus visual $\Delta_{\mathrm{conv}}$ for six omni models, pooling horizons and averaging interfaces; modality subsets differ. (c) MCQ minus OpenQA accuracy by horizon; boxes, lines, and whiskers show the interquartile range, median, and range across 13 models. Details: Appendix~\ref{app:full-results}.}
\label{fig:profiles}
\end{figure}

\begin{samepage}
We hypothesize two complementary demands: accessing required history and binding it to the current request.\par\nopagebreak[4]

\noindent\textbf{(1) History access.} The ability to recover the historical question or media content needed to answer the current request.\par\nopagebreak[4]

\noindent\textbf{(2) Cross-turn binding.} The ability to pair the question with its intended evidence across turns while resisting competing within-turn associations.\par
\end{samepage}

Figure~\ref{fig:profiles}a shows two demand ladders: Single-turn--Reconfirm--Reground and Single-turn--Retrieve--Rebind. Single-turn measures direct answerability; reference conditions introduce historical dependencies; conflict conditions add interference without changing the target. Reground combines an earlier question with current evidence, whereas Rebind combines a current question with earlier evidence. These contrasts motivate the access--binding hypothesis, examined in Section~\ref{sec:diagnosis} through recall probes and interventions on inherited requests and competing evidence.

\subsection{Modality and Interface Expose Distinct Weaknesses}
\label{sec:eval-dimensions}

\textbf{Evidence modality.} Modality sensitivity depends on the operation (Figure~\ref{fig:profiles}b). All six omni-modal models have a larger Rebind $\Delta_{\mathrm{conv}}$ on audio than visual tasks, whereas five have a larger Reground $\Delta_{\mathrm{conv}}$ on visual tasks. This reversal motivates separate diagnosis of the two history roles.

\textbf{Answer interface.} The median absolute OpenQA--MCQ multi-turn difference is 2.5 pp overall but 24.4 pp on Reground. Eleven of 13 configurations score lower on Reground MCQ than OpenQA when pooling horizons; Figure~\ref{fig:profiles}c separates the distributions by horizon. For Qwen3-Omni, overall accuracies are nearly identical (77.26\%/77.28\%), despite Reground scores of 81.76\%/57.76\%. Task-carrying MCQ inherits both question and option mapping, alongside differences in single-turn answerability. Section~\ref{sec:diagnosis-task} tests their restatement separately.

\FloatBarrier
\section{Diagnosing Selective History Failures}
\label{sec:diagnosis}

The aggregate gaps in Section~\ref{sec:evaluation} leave open whether models fail to access useful history or fail to apply it to the intended target. We examine these possibilities through separate probes and controlled interventions. On the task side, we test question recall, then restore parts of the inherited request. On the evidence side, we probe historical source information, then change the current competitor while preserving the target. These comparisons provide behavioral evidence about history use rather than a decomposition of individual errors.

\begin{table}[!t]
\centering
\small
\setlength{\tabcolsep}{3pt}
\renewcommand{\arraystretch}{1.10}
\caption{\textbf{Separating question and option inheritance.} Format-matched MCQ accuracy (\%), $n=128$ per row and condition. Gains (pp) are relative to Neither; omitted components remain in history.}
\label{tab:diagnosis-factorial}
\begin{tabularx}{\textwidth}{@{\hspace{5pt}}ll*{6}{>{\raggedleft\arraybackslash}X}@{\hspace{5pt}}}
\toprule
\multirow{2}{*}[-.6ex]{Model} & \multirow{2}{*}[-.6ex]{Operation} & \multicolumn{4}{c}{Accuracy (\%)} & \multicolumn{2}{c}{Gain over Neither (pp)} \\
\cmidrule(lr){3-6}\cmidrule(l){7-8}
& & Neither & Question & Options & Both & Question & Options \\
\midrule
\multirow{2}{*}{Qwen3-Omni} & Reconfirm & 97.7 & 94.5 & 99.2 & 100.0 & -3.1 & +1.6 \\
 & Reground & 57.8 & 72.7 & 99.2 & 98.4 & +14.8 & \textbf{+41.4} \\
\midrule
\multirow{2}{*}{MiniCPM-o 4.5} & Reconfirm & 85.9 & 90.6 & 98.4 & 97.7 & +4.7 & +12.5 \\
 & Reground & 29.7 & 42.2 & 92.2 & 96.1 & +12.5 & \textbf{+62.5} \\
\bottomrule
\end{tabularx}
\end{table}

\begin{figure}[!t]
\centering
\includegraphics[width=\textwidth]{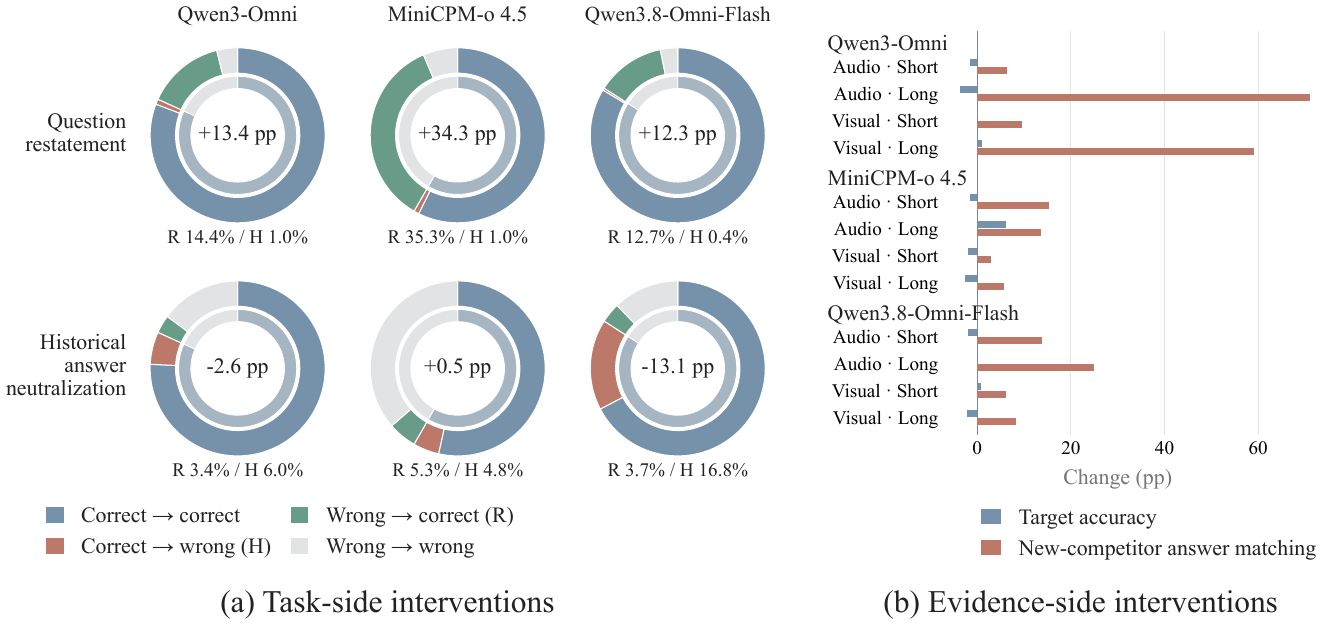}
\caption{\textbf{Task- and evidence-side interventions (OpenQA).} (a) Reground: inner rings show original correctness; outer rings distinguish unchanged, repaired (R), and harmed (H) outcomes. Centers report changes (pp); R/H rates use all pairs. (b) Rebind: changes in target accuracy and replacement-answer matching after a current-media swap. Details: Appendix~\ref{app:diagnostics}.}
\label{fig:diagnosis}
\end{figure}

\subsection{Task-Side: Access and Application}
\label{sec:diagnosis-task}

\textbf{Recalling a question versus applying it.} Qwen3-Omni and Qwen3.8-Omni-Flash recall the inherited question on 96.8\% and 99.8\% of Reground OpenQA probes, yet achieve only 81.8\% and 84.1\% accuracy on the conversational task. Question restatement raises these scores to 95.2\% and 96.4\%. MiniCPM-o 4.5 has weaker recall (60.5\%) and a larger restatement gain (Figure~\ref{fig:diagnosis}a, top). Across these separate probes, high question recall coexists with weaker task performance. Restatement brings accuracy closer to the recall scores, highlighting the value of making an available historical question explicit when applying it to new media.

\begin{table}[!t]
\centering
\caption{\textbf{Two routes to mitigation.} Accuracy (\%) under matched controls. (a) Dev-Val with cached histories ($n=500$ per model/interface). (b) Qwen3-Omni on Test with checkpoint-native histories ($n=5{,}000$ per interface). Single/Multi denote evaluation modes; compare methods within panels.}
\label{tab:mitigation}
\begin{minipage}[t]{.49\textwidth}
\centering
\fontsize{8}{9.5}\selectfont
\setlength{\tabcolsep}{2pt}
\renewcommand{\arraystretch}{1.12}
\begin{tabularx}{\linewidth}{@{}p{52pt}*{4}{>{\centering\arraybackslash}X}@{}}
\toprule
\multirow{2}{*}[-.6ex]{Method} & \multicolumn{2}{c}{Qwen3-Omni} & \multicolumn{2}{c}{MiniCPM-o 4.5}\\
\cmidrule(lr){2-3}\cmidrule(l){4-5}
& OpenQA & MCQ & OpenQA & MCQ\\
\midrule
Neutral & 71.4 & 78.8 & 59.8 & 59.6 \\
Restatement & 76.0 & \textbf{86.4} & 74.2 & \textbf{78.2} \\
Reweighting & 73.2 & 77.6 & 61.4 & 60.2 \\
Routed policy & \textbf{78.2} & 86.0 & \textbf{77.8} & \textbf{78.2} \\
\bottomrule
\end{tabularx}
\par\vspace{3pt}
\normalsize (a) Inference-time assistance
\end{minipage}\hfill
\begin{minipage}[t]{.49\textwidth}
\centering
\fontsize{8}{9.5}\selectfont
\setlength{\tabcolsep}{2pt}
\renewcommand{\arraystretch}{1.12}
\begin{tabularx}{\linewidth}{@{}p{52pt}*{4}{>{\centering\arraybackslash}X}@{}}
\toprule
\multirow{2}{*}[-.6ex]{Method} & \multicolumn{2}{c}{OpenQA} & \multicolumn{2}{c}{MCQ}\\
\cmidrule(lr){2-3}\cmidrule(l){4-5}
& Single & Multi & Single & Multi\\
\midrule
\rule[-4.26pt]{0pt}{14.19pt}Base & 94.60 & 77.26 & 99.12 & 77.28 \\
\rule[-4.26pt]{0pt}{14.19pt}SFT & 94.84 & 80.68 & 99.12 & 79.64 \\
\rule[-4.26pt]{0pt}{14.19pt}SFT+Inv & 94.92 & 80.70 & 99.12 & 79.84 \\
\bottomrule
\end{tabularx}
\par\vspace{3pt}
\normalsize (b) Conversational adaptation
\end{minipage}
\end{table}

\textbf{Restoring the request versus removing an outdated answer.} Question restatement repairs 12.7--35.3\% of valid Reground OpenQA pairs while harming at most 1.0\%. Historical-answer neutralization instead yields inconsistent gains, reducing accuracy for two of the three models (Figure~\ref{fig:diagnosis}a, bottom). Neutralizing an earlier answer therefore does not reproduce the benefit of making the inherited task explicit, limiting a stale-answer-only explanation.

\textbf{Separating question and option inheritance.} MCQ additionally requires inheriting the answer interface. Options-only restatement improves Reground by 41.4/62.5 pp for Qwen3-Omni/MiniCPM-o 4.5, versus 14.8/12.5 pp for questions alone (Table~\ref{tab:diagnosis-factorial}); Reconfirm gains are smaller. Restoring both components adds no gain over options alone for Qwen3-Omni, but further improves MiniCPM-o 4.5. Task inheritance therefore extends beyond question recall to the answer interface, whose options jointly supply candidate answers and their letter mapping.

\FloatBarrier
\subsection{Evidence-Side: Access and Selection}
\label{sec:diagnosis-capture}

\textbf{Accessing historical media.} We first ask a separate source-fact question about the referenced historical media. OpenQA accuracy reaches 66.9--85.7\%, versus 90.4--92.1\% with direct input (Appendix~\ref{app:diagnostics}). These probes show substantial retained source information, potentially supported by historical text as well as media, with higher accuracy under direct input. We next test whether competing current media influences how this information is used for the original request.

\textbf{Changing the competing evidence.} Rebind supplies the complete question. Removing current media improves OpenQA by 39.5/52.9 pp for Qwen3-Omni/MiniCPM-o 4.5, but also reduces input load. We instead replace the current competitor while fixing the historical target. The diagnostic tracks two outcomes: correctness on the unchanged historical target, and agreement with the \emph{same replacement-media answer} before and after the swap. Qwen3-Omni's target accuracy barely changes (52.6\% to 51.7\%), while this matching rate rises from 0.1\% to 36.5\%.

\textbf{Stable accuracy can conceal answer shifts.} Across 12 configurations, target accuracy changes by $-3.6$ to $+6.0$ pp, yet replacement-answer matching increases throughout (Figure~\ref{fig:diagnosis}b). For Qwen3-Omni, this increase is larger in Long (59.1--71.0 pp) than Short (6.5--9.5 pp). An audit restricted to cases where all three media endpoints are answerable alone retains this behavior (Appendix~\ref{app:diagnostics}). This consistent shift under whole-media replacement supports sensitivity to competing content. Across the two sides, the relevant history can remain available while the final answer follows the wrong question--media association. This motivates the two mitigation routes below: making inherited requests explicit and assisting selection of the intended evidence.

\FloatBarrier
\section{Mitigating Selective History Failures}
\label{sec:learning}

The diagnostics motivate assistance with task inheritance and evidence selection. We test inference-time interventions that expose the request or reweight competing context, followed by adaptation on native conversational histories. Table~\ref{tab:mitigation} compares each approach with matched controls; its panels use different splits and history-generation procedures.

\begin{figure}[!t]
\centering
\includegraphics[width=\textwidth]{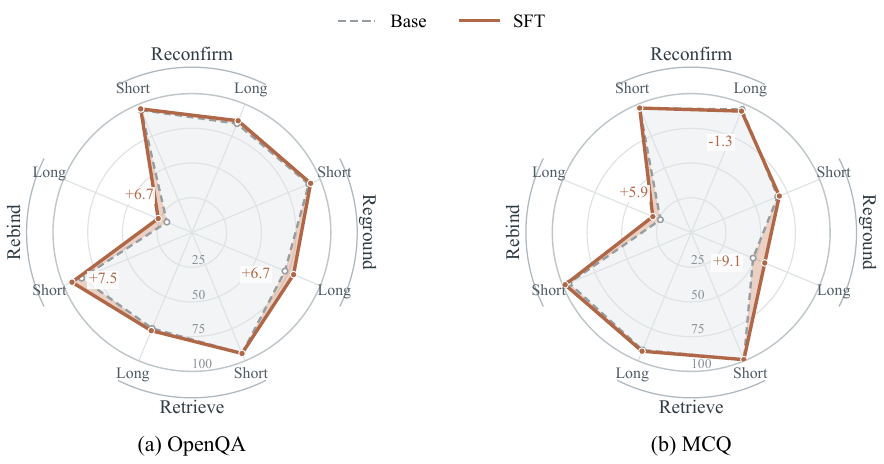}
\caption{\textbf{Capability profiles before and after SFT.} Qwen3-Omni Test multi-turn accuracy (\%), pooled across modalities; all axes span 0--100. Shading highlights changes between contours; selected gains and losses are labeled in pp. Training uses Short OpenQA only.}
\label{fig:training}
\end{figure}

\subsection{Assisting Selection at Inference Time}
\label{sec:learning-inference}

\emph{Restatement} supplies the inherited question and, for MCQ, its original options. \emph{Reweighting} resolves the user-visible media reference and scales attention to non-target media and historical assistant-text keys by 0.5, with renormalization at final-answer positions. A \emph{routed policy} uses restatement for historical-question requests and reweighting for requests containing a complete question. Routing follows the benchmark's explicit media references and request structure, without gold answers or operation labels (Appendix~\ref{app:inference}).

Reweighting yields small, mixed Dev-Val gains: $+1.8/-1.2$ pp on OpenQA/MCQ for Qwen3-Omni and $+1.6/+0.6$ pp for MiniCPM-o 4.5. Routing improves over neutral execution, but its advantage over restatement is mixed across models and interfaces (Table~\ref{tab:mitigation}(a)). The benefit of routing depends on the model and answer interface.

\subsection{Learning from Native Conversational Histories}
\label{sec:learning-setup}

We adapt Qwen3-Omni on 1,018 Short OpenQA views (509 adaptation pairs derived from Dev-Train). SFT+Inv adds a Jensen--Shannon penalty on answer-preserving pairs. Initialization, data order, 204 updates, and auxiliary-forward budgets are matched; each checkpoint generates its own test history (Appendix~\ref{app:adaptation}).

SFT improves multi-turn accuracy by 3.42 pp on OpenQA and 2.36 pp on MCQ while retaining single-turn accuracy (Table~\ref{tab:mitigation}(b)). Long OpenQA improves by 4.36 pp despite Short-only training. Figure~\ref{fig:training} shows improvements on Rebind across both horizons and interfaces, and larger Reground gains on Long conversations. Excluding all input-parent overlap with gradient-training media retains a 3.14 pp OpenQA gain ($n=3,125$; Appendix~\ref{app:adaptation}). The improvement thus extends beyond tasks sharing these training sources.

Residual errors concentrate in difficult conditions: Long Rebind reaches 26.2\% OpenQA and 29.9\% MCQ accuracy after SFT, while Long MCQ Reconfirm declines by 1.3 pp. The OpenQA net gain combines 261 repaired answers with 90 newly incorrect answers (Appendix~\ref{app:adaptation}). These uneven changes identify evidence selection under extended histories as a priority for further improvement.

SFT+Inv adds only 0.02 pp on OpenQA and 0.20 pp on MCQ over matched SFT, with derivation-root sensitivity intervals spanning zero. In this single-model, single-seed study, the observed gains are attributable to SFT, with no clear additional benefit from the consistency term.

\clearpage
\section{Conclusion}
\label{sec:conclusion}

ReTurn provides a controlled evaluation of selective history use, separating the value of historical tasks and evidence from the interference they can introduce. Across omni-modal models, LVLMs, and LALMs, questions that are readily answerable in isolation become substantially harder in conversation. Our diagnostics show that recovering relevant history can coexist with failures to apply it or select the intended source, while mitigation only partially closes these gaps. These findings motivate multimodal systems that preserve task-relevant dependencies and ground each response in the evidence requested by the current turn.

\bibliography{iclr2027_conference}

\begin{thebibliography}{29}
\providecommand{\natexlab}[1]{#1}
\providecommand{\url}[1]{\texttt{#1}}
\expandafter\ifx\csname urlstyle\endcsname\relax
  \providecommand{\doi}[1]{doi: #1}\else
  \providecommand{\doi}{doi: \begingroup \urlstyle{rm}\Url}\fi

\bibitem[{Alibaba Cloud}(2026{\natexlab{a}})]{qwen37plus}
{Alibaba Cloud}.
\newblock {Qwen3.7-Plus}.
\newblock
  \url{https://docs.modelstudio.console.alibabacloud.com/en/model-studio/qwen3-7-plus},
  2026{\natexlab{a}}.
\newblock Alibaba Cloud Model Studio, accessed September 23, 2026.

\bibitem[{Alibaba Cloud}(2026{\natexlab{b}})]{qwen38omniflash}
{Alibaba Cloud}.
\newblock {Qwen3.8-Omni-Flash}.
\newblock
  \url{https://www.alibabacloud.com/help/en/model-studio/qwen3-8-omni-flash},
  2026{\natexlab{b}}.
\newblock Alibaba Cloud Model Studio, accessed September 23, 2026.

\bibitem[Cui et~al.(2026)Cui, Xu, Wang, Yu, Sun, Xu, Wang, He, Ma, Cai, Gui,
  Zhang, Sun, Huang, Chen, Lin, Liu, Gui, Han, Wen, Liu, Wang, Zhang, Wei,
  Chen, Li, Fang, Zhou, Li, Zeng, Xiao, Lin, Han, Sun, Liu, and
  Yao]{DBLP:journals/corr/abs-2604-27393}
Junbo Cui, Bokai Xu, Chongyi Wang, Tianyu Yu, Weiyue Sun, Yingjing Xu, Tianran
  Wang, Zhihui He, Wenshuo Ma, Tianchi Cai, Jiancheng Gui, Luoyuan Zhang, Xian
  Sun, Fuwei Huang, Moye Chen, Zhuo Lin, Hanyu Liu, Qingxin Gui, Qingzhe Han,
  Yuyang Wen, Huiping Liu, Rongkang Wang, Yaqi Zhang, Hongliang Wei, Chi Chen,
  You Li, Kechen Fang, Jie Zhou, Yuxuan Li, Guoyang Zeng, Chaojun Xiao, Yankai
  Lin, Xu~Han, Maosong Sun, Zhiyuan Liu, and Yuan Yao.
\newblock Minicpm-o 4.5: Towards real-time full-duplex omni-modal interaction.
\newblock \emph{CoRR}, abs/2604.27393, 2026.
\newblock \doi{10.48550/ARXIV.2604.27393}.
\newblock URL \url{https://doi.org/10.48550/arXiv.2604.27393}.

\bibitem[Farr\'{e} et~al.(2024)Farr\'{e}, Marafioti, Tunstall, Von~Werra, and
  Wolf]{farre2024finevideo}
Miquel Farr\'{e}, Andi Marafioti, Lewis Tunstall, Leandro Von~Werra, and Thomas
  Wolf.
\newblock {FineVideo}.
\newblock \url{https://huggingface.co/datasets/HuggingFaceFV/finevideo}, 2024.

\bibitem[Ghosh et~al.(2025)Ghosh, Goel, Kim, Kumar, Kong, Lee, Yang,
  Duraiswami, Manocha, Valle, and Catanzaro]{DBLP:conf/nips/GhoshGKKKLYDMVC25}
Sreyan Ghosh, Arushi Goel, Jaehyeon Kim, Sonal Kumar, Zhifeng Kong, Sang{-}gil
  Lee, Chao{-}Han~Huck Yang, Ramani Duraiswami, Dinesh Manocha, Rafael Valle,
  and Bryan Catanzaro.
\newblock Audio flamingo 3: Advancing audio intelligence with fully open large
  audio language models.
\newblock In Danielle Belgrave, Cheng Zhang, Laura~N. Montoya, Hsuan{-}Tien
  Lin, Razvan Pascanu, Piotr Koniusz, Marzyeh Ghassemi, Nancy Chen, Iv{\'{a}}n
  Vladimir~Meza Ru{\'{\i}}z, and Arturo Loaiza{-}Bonilla (eds.), \emph{Advances
  in Neural Information Processing Systems 38: Annual Conference on Neural
  Information Processing Systems 2025, NeurIPS 2025, San Diego, CA, USA,
  December 2-7, 2025 / Mexico City, Mexico, November 30 - December 5, 2025},
  2025.
\newblock URL
  \url{http://papers.nips.cc/paper\_files/paper/2025/hash/3babb6b453cb59d87cb58a1219ef914b-Abstract-Conference.html}.

\bibitem[Gosai et~al.(2026)Gosai, Vuong, Tyagi, Li, You, Bavare, U{\c{c}}ar,
  Fang, Jang, Liu, and He]{DBLP:conf/acl/GosaiVTLYBUFJLH26}
Advait Gosai, Tyler Vuong, Utkarsh Tyagi, Steven Li, Wenjia You, Miheer Bavare,
  Arda U{\c{c}}ar, Zhongwang Fang, Brian Jang, Bing Liu, and Yunzhong He.
\newblock Audio multichallenge: {A} multi-turn evaluation of spoken dialogue
  systems on natural human interaction.
\newblock In Maria Liakata, Viviane~P. Moreira, Jiajun Zhang, and David Jurgens
  (eds.), \emph{Proceedings of the 64th Annual Meeting of the Association for
  Computational Linguistics (Volume 1: Long Papers), {ACL} 2026, San Diego,
  California, United States, July 2-7, 2026}, pp.\  35740--35770. Association
  for Computational Linguistics, 2026.
\newblock \doi{10.18653/V1/2026.ACL-LONG.1654}.
\newblock URL \url{https://doi.org/10.18653/v1/2026.acl-long.1654}.

\bibitem[Hu et~al.(2021)Hu, Shen, Wallis, Allen-Zhu, Li, Wang, Wang, and
  Chen]{hu2021lora}
Edward~J. Hu, Yelong Shen, Phillip Wallis, Zeyuan Allen-Zhu, Yuanzhi Li, Shean
  Wang, Lu~Wang, and Weizhu Chen.
\newblock {LoRA}: Low-rank adaptation of large language models.
\newblock \emph{arXiv preprint arXiv:2106.09685}, 2021.
\newblock URL \url{https://arxiv.org/abs/2106.09685}.

\bibitem[Laban et~al.(2026)Laban, Hayashi, Zhou, and Neville]{laban2026llms}
Philippe Laban, Hiroaki Hayashi, Yingbo Zhou, and Jennifer Neville.
\newblock Llms get lost in multi-turn conversation.
\newblock In \emph{International Conference on Learning Representations},
  volume 2026, pp.\  54738--54778, 2026.

\bibitem[Lee et~al.(2025)Lee, Lee, Zhang, Hwang, Ko, Kim, Yao, Rong, Joo, Han,
  Ko, and Choi]{DBLP:conf/iccv/LeeLZHKKYRJHKC25}
Young{-}Jun Lee, Byung{-}Kwan Lee, Jianshu Zhang, Yechan Hwang, Byungsoo Ko,
  Han{-}Gyu Kim, Dongyu Yao, Xuankun Rong, Eojin Joo, Seung{-}Ho Han, Bowon Ko,
  and Ho{-}Jin Choi.
\newblock Multiverse: {A} multi-turn conversation benchmark for evaluating
  large vision and language models.
\newblock In \emph{{IEEE/CVF} International Conference on Computer Vision,
  {ICCV} 2025, Honolulu, HI, USA, October 19-25, 2025}, pp.\  708--719. {IEEE},
  2025.
\newblock \doi{10.1109/ICCV51701.2025.00074}.
\newblock URL \url{https://doi.org/10.1109/ICCV51701.2025.00074}.

\bibitem[Lin et~al.(2026)Lin, Zhao, Zhang, Wu, Yan, Geng, Wu, Yi, Feng, Zhu,
  et~al.]{lin2026stepaudio}
Bin Lin, Bo~Zhao, Boyang Zhang, Boyong Wu, Chao Yan, Chen Geng, Chen Wu, Cheng
  Yi, Chengli Feng, Chenglin Zhu, et~al.
\newblock {StepAudio 3 Realtime} technical report.
\newblock \emph{arXiv preprint arXiv:2609.14005}, 2026.

\bibitem[Liu et~al.(2024{\natexlab{a}})Liu, Ying, Zhang, Yang, Lin, Zhang, Li,
  Qiao, Luo, Shao, and Zhang]{DBLP:conf/nips/LiuYZYLZL00SZ24}
Shuo Liu, Kaining Ying, Hao Zhang, Yue Yang, Yuqi Lin, Tianle Zhang, Chuanhao
  Li, Yu~Qiao, Ping Luo, Wenqi Shao, and Kaipeng Zhang.
\newblock Convbench: {A} multi-turn conversation evaluation benchmark with
  hierarchical ablation capability for large vision-language models.
\newblock In Amir Globersons, Lester Mackey, Danielle Belgrave, Angela Fan,
  Ulrich Paquet, Jakub~M. Tomczak, and Cheng Zhang (eds.), \emph{Advances in
  Neural Information Processing Systems 37: Annual Conference on Neural
  Information Processing Systems 2024, NeurIPS 2024, Vancouver, BC, Canada,
  December 10 - 15, 2024}, 2024{\natexlab{a}}.
\newblock URL
  \url{http://papers.nips.cc/paper\_files/paper/2024/hash/b69396afc07a9ca3428d194f4db84c02-Abstract-Datasets\_and\_Benchmarks\_Track.html}.

\bibitem[Liu et~al.(2024{\natexlab{b}})Liu, Chu, Zang, Wei, Dong, Zhang, Liang,
  Xiong, Qiao, Lin, and Wang]{DBLP:conf/nips/LiuCZWDZLX0L024}
Ziyu Liu, Tao Chu, Yuhang Zang, Xilin Wei, Xiaoyi Dong, Pan Zhang, Zijian
  Liang, Yuanjun Xiong, Yu~Qiao, Dahua Lin, and Jiaqi Wang.
\newblock {MMDU:} {A} multi-turn multi-image dialog understanding benchmark and
  instruction-tuning dataset for lvlms.
\newblock In Amir Globersons, Lester Mackey, Danielle Belgrave, Angela Fan,
  Ulrich Paquet, Jakub~M. Tomczak, and Cheng Zhang (eds.), \emph{Advances in
  Neural Information Processing Systems 37: Annual Conference on Neural
  Information Processing Systems 2024, NeurIPS 2024, Vancouver, BC, Canada,
  December 10 - 15, 2024}, 2024{\natexlab{b}}.
\newblock URL
  \url{http://papers.nips.cc/paper\_files/paper/2024/hash/1057053100de064a44286239724f7865-Abstract-Datasets\_and\_Benchmarks\_Track.html}.

\bibitem[Loshchilov \& Hutter(2019)Loshchilov and
  Hutter]{loshchilov2019decoupled}
Ilya Loshchilov and Frank Hutter.
\newblock Decoupled weight decay regularization.
\newblock In \emph{International Conference on Learning Representations}, 2019.
\newblock URL \url{https://openreview.net/forum?id=Bkg6RiCqY7}.

\bibitem[Maharana et~al.(2024)Maharana, Lee, Tulyakov, Bansal, Barbieri, and
  Fang]{DBLP:conf/acl/MaharanaLTBBF24}
Adyasha Maharana, Dong{-}Ho Lee, Sergey Tulyakov, Mohit Bansal, Francesco
  Barbieri, and Yuwei Fang.
\newblock Evaluating very long-term conversational memory of {LLM} agents.
\newblock In Lun{-}Wei Ku, Andre Martins, and Vivek Srikumar (eds.),
  \emph{Proceedings of the 62nd Annual Meeting of the Association for
  Computational Linguistics (Volume 1: Long Papers), {ACL} 2024, Bangkok,
  Thailand, August 11-16, 2024}, pp.\  13851--13870. Association for
  Computational Linguistics, 2024.
\newblock \doi{10.18653/V1/2024.ACL-LONG.747}.
\newblock URL \url{https://doi.org/10.18653/v1/2024.acl-long.747}.

\bibitem[NVIDIA(2026)]{DBLP:journals/corr/abs-2604-24954}
NVIDIA.
\newblock Nemotron 3 nano omni: Efficient and open multimodal intelligence.
\newblock \emph{CoRR}, abs/2604.24954, 2026.
\newblock \doi{10.48550/ARXIV.2604.24954}.
\newblock URL \url{https://doi.org/10.48550/arXiv.2604.24954}.

\bibitem[{OpenAI}(2026)]{openai2026gpt56systemcard}
{OpenAI}.
\newblock {GPT-5.6 System Card}.
\newblock OpenAI Deployment Safety Hub, July 2026.
\newblock URL \url{https://deploymentsafety.openai.com/gpt-5-6}.
\newblock Published July 9, 2026.

\bibitem[Pan et~al.(2026)Pan, Xie, Zhang, Wang, Wen, Zhang, Hu, Pan, Huang,
  Gan, Lin, Ping, Li, Wang, Peng, and Liu]{DBLP:conf/acl/PanXZWWZHPHGLPL26}
Yaning Pan, Qianqian Xie, Guohui Zhang, Zekun~Moore Wang, Yongqian Wen,
  Yuanxing Zhang, Haoxuan Hu, Zhiyu Pan, Yibing Huang, Zhidong Gan, Yonghong
  Lin, An~Ping, Shihao Li, Yanghai Wang, Tianhao Peng, and Jiaheng Liu.
\newblock Mt-video-bench: {A} holistic video understanding benchmark for
  evaluating multimodal llms in multi-turn dialogues.
\newblock In Maria Liakata, Viviane~P. Moreira, Jiajun Zhang, and David Jurgens
  (eds.), \emph{Findings of the Association for Computational Linguistics,
  {ACL} 2026, San Diego, California, United States, July 2-7, 2026}, pp.\
  8105--8126. Association for Computational Linguistics, 2026.
\newblock \doi{10.18653/V1/2026.FINDINGS-ACL.397}.
\newblock URL \url{https://doi.org/10.18653/v1/2026.findings-acl.397}.

\bibitem[{Qwen Team}(2025{\natexlab{a}})]{DBLP:journals/corr/abs-2509-17765}
{Qwen Team}.
\newblock Qwen3-omni technical report.
\newblock \emph{CoRR}, abs/2509.17765, 2025{\natexlab{a}}.
\newblock \doi{10.48550/ARXIV.2509.17765}.
\newblock URL \url{https://doi.org/10.48550/arXiv.2509.17765}.

\bibitem[{Qwen Team}(2025{\natexlab{b}})]{DBLP:journals/corr/abs-2511-21631}
{Qwen Team}.
\newblock Qwen3-vl technical report.
\newblock \emph{CoRR}, abs/2511.21631, 2025{\natexlab{b}}.
\newblock \doi{10.48550/ARXIV.2511.21631}.
\newblock URL \url{https://doi.org/10.48550/arXiv.2511.21631}.

\bibitem[Seed(2026)]{seed2026seed2}
Bytedance Seed.
\newblock Seed2. 0 model card: Towards intelligence frontier for real-world
  complexity.
\newblock \emph{arXiv preprint arXiv:2607.00248}, 2026.

\bibitem[Tong et~al.(2025)Tong, Guo, Ran, Chen, Lu, Wang, Li, Zhu, Li, Li,
  et~al.]{tong2025interactiveomni}
Wenwen Tong, Hewei Guo, Dongchuan Ran, Jiangnan Chen, Jiefan Lu, Kaibin Wang,
  Keqiang Li, Xiaoxu Zhu, Jiakui Li, Kehan Li, et~al.
\newblock Interactiveomni: A unified omni-modal model for audio-visual
  multi-turn dialogue.
\newblock \emph{arXiv preprint arXiv:2510.13747}, 2025.

\bibitem[Wang et~al.(2025{\natexlab{a}})Wang, Gao, Gu, Pu, Cui, Wei, Liu, Jing,
  Ye, Shao, Wang, Chen, Zhang, Yang, Wang, Wei, Yin, Li, Cui, Chen, Ding, Tian,
  Wu, Xie, Li, Yang, Duan, Wang, Hou, Hao, Zhang, Li, Zhao, Duan, Deng, Fu, He,
  Wang, He, Shi, He, Xiong, Lv, Wu, Shao, Zhang, Deng, Qi, Ge, Guo, Zhang,
  Zhang, Cao, Lin, Tang, Gao, Huang, Gu, Lyu, Tang, Wang, Lv, Ouyang, Wang,
  Dou, Zhu, Lu, Lin, Dai, Su, Zhou, Chen, Qiao, Wang, and
  Luo]{DBLP:journals/corr/abs-2508-18265}
Weiyun Wang, Zhangwei Gao, Lixin Gu, Hengjun Pu, Long Cui, Xingguang Wei,
  Zhaoyang Liu, Linglin Jing, Shenglong Ye, Jie Shao, Zhaokai Wang, Zhe Chen,
  Hongjie Zhang, Ganlin Yang, Haomin Wang, Qi~Wei, Jinhui Yin, Wenhao Li, Erfei
  Cui, Guanzhou Chen, Zichen Ding, Changyao Tian, Zhenyu Wu, JingJing Xie,
  Zehao Li, Bowen Yang, Yuchen Duan, Xuehui Wang, Zhi Hou, Haoran Hao, Tianyi
  Zhang, Songze Li, Xiangyu Zhao, Haodong Duan, Nianchen Deng, Bin Fu, Yinan
  He, Yi~Wang, Conghui He, Botian Shi, Junjun He, Yingtong Xiong, Han Lv, Lijun
  Wu, Wenqi Shao, Kaipeng Zhang, Huipeng Deng, Biqing Qi, Jiaye Ge, Qipeng Guo,
  Wenwei Zhang, Songyang Zhang, Maosong Cao, Junyao Lin, Kexian Tang, Jianfei
  Gao, Haian Huang, Yuzhe Gu, Chengqi Lyu, Huanze Tang, Rui Wang, Haijun Lv,
  Wanli Ouyang, Limin Wang, Min Dou, Xizhou Zhu, Tong Lu, Dahua Lin, Jifeng
  Dai, Weijie Su, Bowen Zhou, Kai Chen, Yu~Qiao, Wenhai Wang, and Gen Luo.
\newblock Internvl3.5: Advancing open-source multimodal models in versatility,
  reasoning, and efficiency.
\newblock \emph{CoRR}, abs/2508.18265, 2025{\natexlab{a}}.
\newblock \doi{10.48550/ARXIV.2508.18265}.
\newblock URL \url{https://doi.org/10.48550/arXiv.2508.18265}.

\bibitem[Wang et~al.(2025{\natexlab{b}})Wang, Wang, Chen, Wu, Zhao, and
  Zheng]{DBLP:conf/cvpr/WangWCW0Z25}
Yuxuan Wang, Yueqian Wang, Bo~Chen, Tong Wu, Dongyan Zhao, and Zilong Zheng.
\newblock Omnimmi: {A} comprehensive multi-modal interaction benchmark in
  streaming video contexts.
\newblock In \emph{{IEEE/CVF} Conference on Computer Vision and Pattern
  Recognition, {CVPR} 2025, Nashville, TN, USA, June 11-15, 2025}, pp.\
  18925--18935. Computer Vision Foundation / {IEEE}, 2025{\natexlab{b}}.
\newblock \doi{10.1109/CVPR52734.2025.01763}.
\newblock URL
  \url{https://openaccess.thecvf.com/content/CVPR2025/html/Wang\_OmniMMI\_A\_Comprehensive\_Multi-modal\_Interaction\_Benchmark\_in\_Streaming\_Video\_Contexts\_CVPR\_2025\_paper.html}.

\bibitem[Wu et~al.(2025)Wu, Wang, Yu, Zhang, Chang, and
  Yu]{DBLP:conf/iclr/WuWYZCY25}
Di~Wu, Hongwei Wang, Wenhao Yu, Yuwei Zhang, Kai{-}Wei Chang, and Dong Yu.
\newblock Longmemeval: Benchmarking chat assistants on long-term interactive
  memory.
\newblock In \emph{The Thirteenth International Conference on Learning
  Representations, {ICLR} 2025, Singapore, April 24-28, 2025}. OpenReview.net,
  2025.
\newblock URL \url{https://openreview.net/forum?id=pZiyCaVuti}.

\bibitem[Xiaomi(2025)]{DBLP:journals/corr/abs-2512-23808}
LLM{-}Core Xiaomi.
\newblock Mimo-audio: Audio language models are few-shot learners.
\newblock \emph{CoRR}, abs/2512.23808, 2025.
\newblock \doi{10.48550/ARXIV.2512.23808}.
\newblock URL \url{https://doi.org/10.48550/arXiv.2512.23808}.

\bibitem[{XiaomiMiMo}(2026)]{mimov25}
{XiaomiMiMo}.
\newblock {MiMo-V2.5}.
\newblock \url{https://huggingface.co/collections/XiaomiMiMo/mimo-v25}, 2026.

\bibitem[Xue et~al.(2025)Xue, Tang, Hu, Liu, Huang, Li, Liu, Xu, Zhang, Feng,
  Xie, Razzak, Ge, Su, He, and Qiao]{DBLP:conf/acl/XueTHLHLLXZF0RG25}
Haochen Xue, Feilong Tang, Ming Hu, Yexin Liu, Qidong Huang, Yulong Li,
  Chengzhi Liu, Zhongxing Xu, Chong Zhang, Chun{-}Mei Feng, Yutong Xie, Imran
  Razzak, Zongyuan Ge, Jionglong Su, Junjun He, and Yu~Qiao.
\newblock {MMRC:} {A} large-scale benchmark for understanding multimodal large
  language model in real-world conversation.
\newblock In Wanxiang Che, Joyce Nabende, Ekaterina Shutova, and Mohammad~Taher
  Pilehvar (eds.), \emph{Proceedings of the 63rd Annual Meeting of the
  Association for Computational Linguistics (Volume 1: Long Papers), {ACL}
  2025, Vienna, Austria, July 27 - August 1, 2025}, pp.\  22477--22503.
  Association for Computational Linguistics, 2025.
\newblock \doi{10.18653/V1/2025.ACL-LONG.1096}.
\newblock URL \url{https://doi.org/10.18653/v1/2025.acl-long.1096}.

\bibitem[Yu et~al.(2026)Yu, Wang, Wang, Huang, Ma, He, Cai, Chen, Huang, Zhao,
  et~al.]{yu2026minicpm}
Tianyu Yu, Zefan Wang, Chongyi Wang, Fuwei Huang, Wenshuo Ma, Zhihui He,
  Tianchi Cai, Weize Chen, Yuxiang Huang, Ranchi Zhao, et~al.
\newblock {MiniCPM-V 4.5}: Cooking efficient {MLLMs} via architecture, data,
  and training recipe.
\newblock In \emph{Proceedings of the IEEE/CVF Conference on Computer Vision
  and Pattern Recognition}, pp.\  11704--11715, 2026.

\bibitem[Zhang et~al.(2024)Zhang, Wu, Li, Li, Ma, Liu, and
  Li]{zhang2024videoinstructiontuningsynthetic}
Yuanhan Zhang, Jinming Wu, Wei Li, Bo~Li, Zejun Ma, Ziwei Liu, and Chunyuan Li.
\newblock {LLaVA-Video}: Video instruction tuning with synthetic data.
\newblock \emph{arXiv preprint arXiv:2410.02713}, 2024.
\newblock URL \url{https://arxiv.org/abs/2410.02713}.

\end{thebibliography}
\bibliographystyle{iclr2027_conference}
\clearpage
\appendix
\clubpenalty=10000
\widowpenalty=10000
\displaywidowpenalty=10000
\makeatletter
\setlength{\@fptop}{0pt}
\setlength{\@fpsep}{24pt}
\setlength{\@fpbot}{0pt plus 1fil}
\makeatother
\raggedbottom
\setcounter{topnumber}{4}
\setcounter{bottomnumber}{3}
\renewcommand{\topfraction}{.95}
\renewcommand{\bottomfraction}{.85}
\renewcommand{\textfraction}{.05}
\renewcommand{\floatpagefraction}{.75}
\setcounter{totalnumber}{6}

\section{Construction, Provenance, and Exposure}
\label{app:provenance}

\paragraph{Evidence before dialogue construction.}
ReTurn uses traceable clips from FineVideo and LLaVA-Video-178K, including previously audited assets with retained provenance. Source annotations and captions support localization; they are not themselves the answer evidence. Newly introduced clips or changed evidence requirements are checked against actual audio/video with Doubao-Seed-2.0-Lite~\citep{seed2026seed2}. GPT-5.6 Sol (\texttt{gpt-5.6-sol})~\citep{openai2026gpt56systemcard} assists question analysis, task-consistency checks, and ambiguous cases, with visual-frame inspection where needed. Text-only review does not substitute for audio verification. Each base question records its parent video, temporal window, required modality, answer granularity, and supporting evidence. Verification outcomes are reused only for identical inputs and evidence requirements. Verification is automated; human annotation agreement was not measured.

\paragraph{Two target-preserving constructions.}
Reconfirm and Reground share the current target media, initial complete question, final follow-up, gold answer, options and order, and intervening user media and questions. The historical media supports either the same or a different answer to that question. The model produces its own historical replies, so changing historical media can also change subsequent assistant text. Retrieve and Rebind instead share the historical target, its position and preceding prefix, final complete question, gold answer, and options; their current non-target media provides weaker or stronger answer competition. Both conditions contain current media. The historical question is constructed separately from the final question to reduce direct answer copying. These are two local matched constructions, not a complete factorial crossing of independent task and evidence factors. Correctness on the agreeing Reconfirm control alone does not establish that the model re-inspected current media: the historical answer can already match the target.

\paragraph{Instantiation and partitioning.}
The benchmark contains 7,000 base tasks in 3,500 matched pairs: 1,750 tasks per operation, 3,500 per required modality, and 3,500 per horizon. Short and Long contain two and five media-bearing user turns, respectively, including the final turn. Long task-carrying histories insert three intervening media question/answer turns; evidence-carrying targets can occupy different historical positions. Short and Long are not universally item-matched. Each task has OpenQA and four-option MCQ views, yielding 14,000 task/interface views rather than two independent datasets. Modality labels identify necessary answer evidence, not necessarily the modalities presented to an omni-modal model.

Development calibration uses Qwen3-Omni single-turn and native multi-turn runs, supplemented by selective checks on other models, to guide refinement and expansion toward high single-turn answerability and measurable history effects. Small development batches establish questions, gold answers, pairing constraints, and scoring granularity; low single-turn performance prompts checks of evidence, wording, input consumption, and truncation. The approximate 80\% single-turn target was a development-set observation goal, not an item-level acceptance threshold. Neither universal single-turn correctness nor large degradation in every condition was required. The final split uses metadata grouping for matched pairs, recorded fact/derivation roots, and shared requests, without reading final model accuracies: 2,000 Dev tasks, comprising 1,500 Dev-Train and 500 Dev-Val tasks, plus 5,000 Test tasks. Different facts can still share parent media. Of the 5,000 Test tasks, 1,028 previously belonged to TRAIN/DEV partitions; 2,268 carry broader component-level development tags, with 536 tasks in both groups. These indicators describe different exposure histories, not gradient-training counts. No exact adaptation task ID occurs in Test, but repartitioning does not remove prior research exposure.

\begin{table}[!htbp]
\centering
\small
\setlength{\tabcolsep}{4pt}
\renewcommand{\arraystretch}{1.05}
\caption{\textbf{Actual-input source and development exposure.} Counts are test tasks satisfying each indicator, not numbers of distinct videos; indicators overlap.}
\label{tab:app-exposure}
\begin{tabular*}{\textwidth}{@{\extracolsep{\fill}}lr@{}}
\toprule
Exposure indicator & Test tasks ($n$) \\
\midrule
Any input parent occurs in development & 3,072 \\
Any input parent occurs in gradient-used data & 1,875 \\
An exact input window occurs in development & 2,947 \\
An exact input window occurs in gradient-used data & 1,571 \\
Task has an earlier training/development tag & 2,268 \\
Target parent occurs in development & 918 \\
Target parent occurs in gradient-used data & 486 \\
\bottomrule
\end{tabular*}
\end{table}

\paragraph{Test-set source exposure.}
Table~\ref{tab:app-exposure} audits the final 5,000 Test tasks using actual model inputs, excluding unused partner/donor metadata. Gradient exposure refers to the 750 unique task IDs used for adaptation (Appendix~\ref{app:adaptation}), rather than the entire Dev-Train partition. Indicators overlap and count test tasks rather than unique parents. Target-parent disjointness concerns target media; other inputs may still share development sources. Shared media and prior development exposure also preclude interpreting task-level sample size as the number of independent source observations. Requiring every input parent to be disjoint from Dev retains 1,928 tasks but removes all Long Retrieve/Rebind tasks. Source-filtered aggregates describe performance under different exposure and task-composition profiles.

Code, preview tasks, and the full benchmark will be made available through the project repository, with media distributed under their source licenses and access terms.

\section{Models and Evaluation Protocol}
\label{app:models}

\paragraph{Native histories and media scope.}
Each model generates its own assistant replies recursively from the chronological prefix, separately for OpenQA and MCQ. Earlier errors remain in the history; gold replies are not substituted. Exact successful requests may be reused, but responses to a different wording or prefix are not interchangeable. LVLMs consume visual-only media and LALMs consume the original companion audio without an ASR substitution. The two hosted omni-modal configurations explicitly use full synchronized audio/video for both visual-evidence and audio-evidence tasks. This input policy does not make every task a joint audio-visual reasoning task.

\begin{table}[!htbp]
\centering
\small
\setlength{\tabcolsep}{4pt}
\renewcommand{\arraystretch}{1.12}
\caption{\textbf{Evaluated model configurations.} References identify model reports or official documentation; identifiers specify evaluated configurations. A/V denote audio/visual evidence subsets. Limits are configured output-token limits; arrows indicate a length-retry cap where recorded. Sampling and media parameters below are configuration-specific.}
\label{tab:app-models}
\begin{tabular}{@{}>{\raggedright\arraybackslash}p{0.28\textwidth}>{\raggedright\arraybackslash}p{0.12\textwidth}>{\raggedright\arraybackslash}p{0.54\textwidth}@{}}
\toprule
Display name & Scope; limit & Execution and recorded settings \\
\midrule
Qwen3.8-Omni-Flash\newline\citep{qwen38omniflash} & A/V; 1,024 $\to$ 4,096 & Hosted \texttt{qwen3.8-omni-flash}; temperature 0, reasoning effort none, 2 fps; companion WAV and silent video. \\
Qwen3-Omni\newline\citep{DBLP:journals/corr/abs-2509-17765} & A/V; 128 & Checkpoint: {\small\path{Qwen3-Omni-30B-A3B-Instruct}}. Local native runner; frozen formal decoding configuration. \\
Doubao-Seed-2.0-Lite\newline\citep{seed2026seed2} & A/V; 1,024 $\to$ 4,096 & Hosted \path{doubao-seed-2-0-lite-260428}; temperature 0, 2 fps; audio, silent video, and text in order. \\
MiniCPM-o 4.5\newline\citep{DBLP:journals/corr/abs-2604-27393} & A/V; 256 & Recorded runner key: {\small\path{minicpmo_45_instruct}}. Local native runner; frozen formal decoding configuration. \\
MiMo-V2.5\newline\citep{mimov25} & A/V; 4,096 & Recorded runner key: {\small\path{mimo_v2_5}}. Local SGLang, tensor parallelism 4; configured adaptive cap 8,192. \\
Nemotron-3-Nano-Omni\newline\citep{DBLP:journals/corr/abs-2604-24954} & A/V; 1,024 & Model: {\small\path{Nemotron-3-Nano-Omni-30B}}. Local SGLang single-turn and vLLM multi-turn execution. \\
\midrule
Qwen3.7-Plus\newline\citep{qwen37plus} & V; 1,024 & Hosted \texttt{qwen3.7-plus}; temperature 0, 2 fps, video only, without added clip labels. \\
Qwen3-VL\newline\citep{DBLP:journals/corr/abs-2511-21631} & V; 256 $\to$ 1,024 & Checkpoint: {\small\path{Qwen/Qwen3-VL-30B-A3B-Instruct}}. Official local Hugging Face implementation; bfloat16, SDPA. \\
MiniCPM-V 4.5\newline\citep{yu2026minicpm} & V; 256 $\to$ 1,024 & Checkpoint: {\small\path{openbmb/MiniCPM-V-4_5}}. Official local implementation; bfloat16, SDPA. \\
InternVL3.5\newline\citep{DBLP:journals/corr/abs-2508-18265} & V; 256 $\to$ 1,024 & Checkpoint: {\small\path{OpenGVLab/InternVL3_5-30B-A3B}}. Official remote-code implementation; bfloat16, FlashAttention 2 when available. \\
\midrule
StepAudio 3\newline\citep{lin2026stepaudio} & A; 1,024 & Chat API \href{https://platform.stepfun.ai/docs/en/guides/models/stepaudio-3-realtime}{\texttt{stepaudio-3-chat-preview}}; temperature 0; original companion WAV. \\
Audio Flamingo 3\newline\citep{DBLP:conf/nips/GhoshGKKKLYDMVC25} & A; 256 $\to$ 1,024 & Recorded runner key: {\small\path{audio_flamingo3_hf}}. Official local implementation; SDPA and bfloat16 autocast. \\
MiMo-Audio\newline\citep{DBLP:journals/corr/abs-2512-23808} & A; 256 & Model: {\small\path{MiMo-Audio-7B-Instruct}}. Local implementation; original companion WAV and frozen turn boundaries. \\
\bottomrule
\end{tabular}
\end{table}

\paragraph{Runtime details.}
Table~\ref{tab:app-models} identifies the evaluated configurations. Visual-only local models use silent video sampled at 2 fps from frame zero, with maximum dimension 768 and native temporal padding. Qwen3-VL, MiniCPM-V 4.5, InternVL3.5, and Audio Flamingo 3 use greedy decoding, a 256-token limit (1,024 on retry), and batch sizes 2/4/1/8, respectively. Qwen3-VL and Audio Flamingo 3 use Transformers 5.12.1; MiniCPM-V 4.5 and InternVL3.5 use 4.51.2; all use PyTorch 2.12.1. MiniCPM-V 4.5 disables image IDs and thinking, using one slice and temporal packing 1. Audio Flamingo 3 retains official FP32 positional embeddings under bfloat16 autocast. MiMo-Audio uses Python 3.10.20, PyTorch 2.6.0, Transformers 4.49.0, and FlashAttention 2.7.4.post1. Local omni-modal models retain their recorded decoding configurations rather than hosted-model temperature settings. Nemotron-3-Nano-Omni uses SGLang for single-turn and vLLM for multi-turn execution, so its gap may include backend effects. Cross-model comparisons also reflect differences in input processing and decoding.

\paragraph{Single-turn reference and valid responses.}
The single-turn input supplies the target media and complete question: current media for task-carrying history, historical target media for evidence-carrying history. It is an empirical answerability reference, not a mathematical ceiling or simply the multi-turn prompt with history deleted. Scores use task views with valid responses in both conditions. The 95,000 expected model/task/interface views yield 94,984 valid pairs: Qwen3.8-Omni-Flash excludes 9, InternVL3.5 excludes 1, and StepAudio 3 excludes 6. Other model views are complete. The precise affected operation/horizon cells are listed beneath the full result tables.

\FloatBarrier
\section{Full Interface-Specific Results and Aggregation}
\label{app:full-results}

Tables~\ref{tab:app-full-openqa-short}--\ref{tab:app-full-mcq-long} report Test single-turn (ST) and multi-turn (MT) accuracy on $n$ common valid response pairs per cell.

\begin{table}[!htbp]
\centering
\fontsize{8.5}{10}\selectfont
\setlength{\tabcolsep}{1.5pt}
\renewcommand{\arraystretch}{1.18}
\caption{\textbf{OpenQA, Short: full Test results.} Single-turn (ST) and multi-turn (MT) accuracy (\%); $\Delta_{\mathrm{conv}}=\mathrm{ST}-\mathrm{MT}$ (pp).}
\label{tab:app-full-openqa-short}
\begin{tabular*}{\textwidth}{@{\extracolsep{\fill}}lrrrrrrrrrrrr@{}}
\toprule
 & \multicolumn{6}{c}{Task-carrying history} & \multicolumn{6}{c}{Evidence-carrying history} \\
\cmidrule(lr){2-7}\cmidrule(l){8-13}
\multirow{2}{*}{Model} & \multicolumn{3}{c}{Reconfirm} & \multicolumn{3}{c}{Reground} & \multicolumn{3}{c}{Retrieve} & \multicolumn{3}{c}{Rebind} \\
\cmidrule(lr){2-4}\cmidrule(lr){5-7}\cmidrule(lr){8-10}\cmidrule(l){11-13}
 & ST & MT & $\Delta_{\mathrm{conv}}$ & ST & MT & $\Delta_{\mathrm{conv}}$ & ST & MT & $\Delta_{\mathrm{conv}}$ & ST & MT & $\Delta_{\mathrm{conv}}$ \\
\midrule
\rowcolor{FamilyGray}\multicolumn{13}{l}{\textit{Omni-modal}} \\
Qwen3.8-Omni-Flash & 96.2 & 94.4 & 1.8 & 96.2 & 95.0 & 1.1 & 93.0 & 94.4 & -1.4 & 92.9 & 80.6 & 12.3 \\
Qwen3-Omni & 95.0 & 95.4 & -0.3 & 95.0 & 91.0 & 4.0 & 94.2 & 94.2 & 0.0 & 94.2 & 85.9 & 8.3 \\
Doubao-Seed-2.0-Lite & 97.4 & 95.4 & 2.1 & 97.4 & 93.1 & 4.3 & 95.4 & 94.7 & 0.6 & 95.4 & 75.8 & 19.5 \\
MiniCPM-o 4.5 & 93.9 & 88.5 & 5.4 & 93.9 & 80.8 & 13.1 & 92.3 & 91.0 & 1.3 & 92.3 & 54.4 & 37.9 \\
MiMo-V2.5 & 92.6 & 90.1 & 2.6 & 92.6 & 79.5 & 13.1 & 90.2 & 87.7 & 2.6 & 90.2 & 63.5 & 26.7 \\
Nemotron-3-Nano-Omni & 90.1 & 92.5 & -2.4 & 90.1 & 75.4 & 14.7 & 92.0 & 93.4 & -1.4 & 92.0 & 82.2 & 9.8 \\
\rowcolor{FamilyGray}\multicolumn{13}{l}{\textit{LVLM}} \\
Qwen3.7-Plus & 94.9 & 94.6 & 0.3 & 94.9 & 92.0 & 2.9 & 92.0 & 93.3 & -1.3 & 92.0 & 90.4 & 1.6 \\
Qwen3-VL & 92.0 & 90.7 & 1.3 & 92.0 & 87.8 & 4.2 & 90.1 & 90.7 & -0.6 & 90.1 & 82.4 & 7.7 \\
MiniCPM-V 4.5 & 87.5 & 88.8 & -1.3 & 87.5 & 60.9 & 26.6 & 90.7 & 91.7 & -1.0 & 90.7 & 87.9 & 2.9 \\
InternVL3.5 & 80.8 & 42.9 & 37.8 & 80.8 & 29.8 & 51.0 & 89.5 & 87.5 & 1.9 & 89.4 & 65.1 & 24.4 \\
\rowcolor{FamilyGray}\multicolumn{13}{l}{\textit{LALM}} \\
StepAudio 3 & 98.4 & 99.0 & -0.6 & 98.4 & 99.4 & -1.0 & 99.0 & 87.5 & 11.6 & 99.0 & 34.1 & 65.0 \\
Audio Flamingo 3 & 97.8 & 95.5 & 2.2 & 97.8 & 47.3 & 50.5 & 93.9 & 93.3 & 0.6 & 93.9 & 54.5 & 39.4 \\
MiMo-Audio & 96.2 & 86.3 & 9.9 & 96.2 & 72.5 & 23.6 & 93.6 & 73.1 & 20.5 & 93.6 & 18.9 & 74.7 \\
\bottomrule
\end{tabular*}
\par\smallskip
\begin{minipage}{\textwidth}\footnotesize
Valid $n$ per operation: omni-modal 625; LVLM 312/313 and LALM 313/312 for task-/evidence-carrying history. Exceptions: InternVL3.5 Rebind 312; Qwen3.8-Omni-Flash Rebind 624; StepAudio 3 Rebind 311; StepAudio 3 Retrieve 311.
\end{minipage}
\end{table}

\begin{table}[!htbp]
\centering
\fontsize{8.5}{10}\selectfont
\setlength{\tabcolsep}{1.5pt}
\renewcommand{\arraystretch}{1.18}
\caption{\textbf{OpenQA, Long: full Test results.} Single-turn (ST) and multi-turn (MT) accuracy (\%); $\Delta_{\mathrm{conv}}=\mathrm{ST}-\mathrm{MT}$ (pp).}
\label{tab:app-full-openqa-long}
\begin{tabular*}{\textwidth}{@{\extracolsep{\fill}}lrrrrrrrrrrrr@{}}
\toprule
 & \multicolumn{6}{c}{Task-carrying history} & \multicolumn{6}{c}{Evidence-carrying history} \\
\cmidrule(lr){2-7}\cmidrule(l){8-13}
\multirow{2}{*}{Model} & \multicolumn{3}{c}{Reconfirm} & \multicolumn{3}{c}{Reground} & \multicolumn{3}{c}{Retrieve} & \multicolumn{3}{c}{Rebind} \\
\cmidrule(lr){2-4}\cmidrule(lr){5-7}\cmidrule(lr){8-10}\cmidrule(l){11-13}
 & ST & MT & $\Delta_{\mathrm{conv}}$ & ST & MT & $\Delta_{\mathrm{conv}}$ & ST & MT & $\Delta_{\mathrm{conv}}$ & ST & MT & $\Delta_{\mathrm{conv}}$ \\
\midrule
\rowcolor{FamilyGray}\multicolumn{13}{l}{\textit{Omni-modal}} \\
Qwen3.8-Omni-Flash & 96.9 & 91.3 & 5.6 & 97.0 & 73.1 & 23.9 & 93.0 & 93.8 & -0.8 & 93.0 & 74.6 & 18.4 \\
Qwen3-Omni & 94.9 & 84.8 & 10.1 & 94.9 & 72.5 & 22.4 & 94.2 & 74.7 & 19.5 & 94.2 & 19.5 & 74.7 \\
Doubao-Seed-2.0-Lite & 97.8 & 82.9 & 14.9 & 97.8 & 77.9 & 19.8 & 95.4 & 76.5 & 18.9 & 95.4 & 14.7 & 80.6 \\
MiniCPM-o 4.5 & 93.8 & 40.0 & 53.8 & 93.8 & 35.8 & 57.9 & 92.3 & 78.9 & 13.4 & 92.3 & 15.5 & 76.8 \\
MiMo-V2.5 & 93.8 & 59.7 & 34.1 & 93.8 & 48.5 & 45.3 & 90.2 & 65.6 & 24.6 & 90.2 & 15.4 & 74.9 \\
Nemotron-3-Nano-Omni & 90.6 & 34.7 & 55.8 & 90.6 & 25.8 & 64.8 & 92.0 & 81.8 & 10.2 & 92.0 & 19.7 & 72.3 \\
\rowcolor{FamilyGray}\multicolumn{13}{l}{\textit{LVLM}} \\
Qwen3.7-Plus & 95.8 & 89.4 & 6.4 & 95.8 & 64.1 & 31.7 & 92.0 & 92.7 & -0.6 & 92.0 & 79.2 & 12.8 \\
Qwen3-VL & 92.3 & 80.1 & 12.2 & 92.3 & 63.8 & 28.5 & 90.1 & 60.1 & 30.0 & 90.1 & 23.0 & 67.1 \\
MiniCPM-V 4.5 & 87.5 & 79.5 & 8.0 & 87.5 & 28.8 & 58.7 & 90.7 & 82.4 & 8.3 & 90.7 & 68.4 & 22.4 \\
InternVL3.5 & 81.1 & 8.7 & 72.4 & 81.1 & 7.7 & 73.4 & 89.5 & 73.8 & 15.7 & 89.5 & 37.1 & 52.4 \\
\rowcolor{FamilyGray}\multicolumn{13}{l}{\textit{LALM}} \\
StepAudio 3 & 98.7 & 98.7 & 0.0 & 98.7 & 99.0 & -0.3 & 99.0 & 91.0 & 8.0 & 99.0 & 51.9 & 47.1 \\
Audio Flamingo 3 & 97.8 & 64.2 & 33.5 & 97.8 & 45.7 & 52.1 & 93.9 & 86.9 & 7.1 & 93.9 & 19.9 & 74.0 \\
MiMo-Audio & 97.4 & 50.8 & 46.6 & 97.4 & 42.2 & 55.3 & 93.6 & 52.2 & 41.3 & 93.6 & 4.2 & 89.4 \\
\bottomrule
\end{tabular*}
\par\smallskip
\begin{minipage}{\textwidth}\footnotesize
Valid $n$ per operation: omni-modal 625; LVLM 312/313 and LALM 313/312 for task-/evidence-carrying history. Exceptions: Qwen3.8-Omni-Flash Reconfirm 621; Qwen3.8-Omni-Flash Reground 624.
\end{minipage}
\end{table}

\begin{table}[!htbp]
\centering
\fontsize{8.5}{10}\selectfont
\setlength{\tabcolsep}{1.5pt}
\renewcommand{\arraystretch}{1.18}
\caption{\textbf{MCQ, Short: full Test results.} Single-turn (ST) and multi-turn (MT) accuracy (\%); $\Delta_{\mathrm{conv}}=\mathrm{ST}-\mathrm{MT}$ (pp).}
\label{tab:app-full-mcq-short}
\begin{tabular*}{\textwidth}{@{\extracolsep{\fill}}lrrrrrrrrrrrr@{}}
\toprule
 & \multicolumn{6}{c}{Task-carrying history} & \multicolumn{6}{c}{Evidence-carrying history} \\
\cmidrule(lr){2-7}\cmidrule(l){8-13}
\multirow{2}{*}{Model} & \multicolumn{3}{c}{Reconfirm} & \multicolumn{3}{c}{Reground} & \multicolumn{3}{c}{Retrieve} & \multicolumn{3}{c}{Rebind} \\
\cmidrule(lr){2-4}\cmidrule(lr){5-7}\cmidrule(lr){8-10}\cmidrule(l){11-13}
 & ST & MT & $\Delta_{\mathrm{conv}}$ & ST & MT & $\Delta_{\mathrm{conv}}$ & ST & MT & $\Delta_{\mathrm{conv}}$ & ST & MT & $\Delta_{\mathrm{conv}}$ \\
\midrule
\rowcolor{FamilyGray}\multicolumn{13}{l}{\textit{Omni-modal}} \\
Qwen3.8-Omni-Flash & 98.6 & 98.6 & 0.0 & 98.6 & 93.9 & 4.6 & 99.0 & 99.0 & 0.0 & 99.0 & 81.9 & 17.1 \\
Qwen3-Omni & 98.6 & 97.0 & 1.6 & 98.6 & 67.4 & 31.2 & 99.5 & 99.2 & 0.3 & 99.5 & 95.2 & 4.3 \\
Doubao-Seed-2.0-Lite & 99.4 & 98.7 & 0.6 & 99.4 & 71.8 & 27.5 & 99.5 & 98.9 & 0.6 & 99.5 & 86.7 & 12.8 \\
MiniCPM-o 4.5 & 97.3 & 96.0 & 1.3 & 97.3 & 32.3 & 65.0 & 97.9 & 96.5 & 1.4 & 97.9 & 49.0 & 49.0 \\
MiMo-V2.5 & 97.8 & 89.0 & 8.8 & 97.8 & 46.9 & 50.9 & 97.4 & 96.3 & 1.1 & 97.4 & 71.4 & 26.1 \\
Nemotron-3-Nano-Omni & 97.9 & 99.0 & -1.1 & 97.9 & 34.4 & 63.5 & 97.4 & 99.4 & -1.9 & 97.4 & 82.1 & 15.4 \\
\rowcolor{FamilyGray}\multicolumn{13}{l}{\textit{LVLM}} \\
Qwen3.7-Plus & 98.7 & 98.4 & 0.3 & 98.7 & 80.1 & 18.6 & 97.8 & 98.1 & -0.3 & 97.8 & 97.4 & 0.3 \\
Qwen3-VL & 97.1 & 98.7 & -1.6 & 97.1 & 32.1 & 65.1 & 98.7 & 98.4 & 0.3 & 98.7 & 92.0 & 6.7 \\
MiniCPM-V 4.5 & 96.5 & 94.2 & 2.2 & 96.5 & 20.8 & 75.6 & 97.8 & 97.8 & 0.0 & 97.8 & 92.7 & 5.1 \\
InternVL3.5 & 94.2 & 93.3 & 1.0 & 94.2 & 11.9 & 82.4 & 95.8 & 97.8 & -1.9 & 95.8 & 78.9 & 16.9 \\
\rowcolor{FamilyGray}\multicolumn{13}{l}{\textit{LALM}} \\
StepAudio 3 & 100.0 & 100.0 & 0.0 & 100.0 & 100.0 & 0.0 & 100.0 & 99.7 & 0.3 & 100.0 & 39.5 & 60.5 \\
Audio Flamingo 3 & 99.7 & 100.0 & -0.3 & 99.7 & 6.4 & 93.3 & 100.0 & 99.4 & 0.6 & 100.0 & 57.7 & 42.3 \\
MiMo-Audio & 98.7 & 97.4 & 1.3 & 98.7 & 13.1 & 85.6 & 96.8 & 66.0 & 30.8 & 96.8 & 21.8 & 75.0 \\
\bottomrule
\end{tabular*}
\par\smallskip
\begin{minipage}{\textwidth}\footnotesize
Valid $n$ per operation: omni-modal 625; LVLM 312/313 and LALM 313/312 for task-/evidence-carrying history. Exceptions: StepAudio 3 Rebind 311; StepAudio 3 Reconfirm 312; StepAudio 3 Reground 312; StepAudio 3 Retrieve 311.
\end{minipage}
\end{table}

\begin{table}[!htbp]
\centering
\fontsize{8.5}{10}\selectfont
\setlength{\tabcolsep}{1.5pt}
\renewcommand{\arraystretch}{1.18}
\caption{\textbf{MCQ, Long: full Test results.} Single-turn (ST) and multi-turn (MT) accuracy (\%); $\Delta_{\mathrm{conv}}=\mathrm{ST}-\mathrm{MT}$ (pp).}
\label{tab:app-full-mcq-long}
\begin{tabular*}{\textwidth}{@{\extracolsep{\fill}}lrrrrrrrrrrrr@{}}
\toprule
 & \multicolumn{6}{c}{Task-carrying history} & \multicolumn{6}{c}{Evidence-carrying history} \\
\cmidrule(lr){2-7}\cmidrule(l){8-13}
\multirow{2}{*}{Model} & \multicolumn{3}{c}{Reconfirm} & \multicolumn{3}{c}{Reground} & \multicolumn{3}{c}{Retrieve} & \multicolumn{3}{c}{Rebind} \\
\cmidrule(lr){2-4}\cmidrule(lr){5-7}\cmidrule(lr){8-10}\cmidrule(l){11-13}
 & ST & MT & $\Delta_{\mathrm{conv}}$ & ST & MT & $\Delta_{\mathrm{conv}}$ & ST & MT & $\Delta_{\mathrm{conv}}$ & ST & MT & $\Delta_{\mathrm{conv}}$ \\
\midrule
\rowcolor{FamilyGray}\multicolumn{13}{l}{\textit{Omni-modal}} \\
Qwen3.8-Omni-Flash & 98.6 & 95.3 & 3.2 & 98.6 & 89.9 & 8.7 & 99.0 & 98.9 & 0.2 & 99.0 & 76.8 & 22.2 \\
Qwen3-Omni & 98.9 & 95.8 & 3.0 & 98.9 & 48.2 & 50.7 & 99.5 & 91.5 & 8.0 & 99.5 & 24.0 & 75.5 \\
Doubao-Seed-2.0-Lite & 99.2 & 82.2 & 17.0 & 99.2 & 62.6 & 36.6 & 99.5 & 91.2 & 8.3 & 99.5 & 29.1 & 70.4 \\
MiniCPM-o 4.5 & 97.3 & 64.6 & 32.6 & 97.3 & 23.7 & 73.6 & 97.9 & 89.0 & 9.0 & 97.9 & 13.8 & 84.2 \\
MiMo-V2.5 & 97.3 & 71.7 & 25.6 & 97.3 & 31.8 & 65.4 & 97.4 & 84.8 & 12.6 & 97.4 & 24.3 & 73.1 \\
Nemotron-3-Nano-Omni & 97.9 & 97.9 & 0.0 & 97.9 & 17.9 & 80.0 & 97.4 & 93.8 & 3.7 & 97.4 & 25.0 & 72.5 \\
\rowcolor{FamilyGray}\multicolumn{13}{l}{\textit{LVLM}} \\
Qwen3.7-Plus & 98.4 & 95.2 & 3.2 & 98.4 & 54.8 & 43.6 & 97.8 & 97.4 & 0.3 & 97.8 & 87.2 & 10.5 \\
Qwen3-VL & 96.5 & 93.9 & 2.6 & 96.5 & 32.1 & 64.4 & 98.7 & 84.3 & 14.4 & 98.7 & 27.8 & 70.9 \\
MiniCPM-V 4.5 & 95.5 & 93.3 & 2.2 & 95.5 & 8.0 & 87.5 & 97.8 & 91.1 & 6.7 & 97.8 & 65.8 & 31.9 \\
InternVL3.5 & 94.6 & 92.9 & 1.6 & 94.6 & 12.2 & 82.4 & 95.8 & 88.5 & 7.3 & 95.8 & 39.6 & 56.2 \\
\rowcolor{FamilyGray}\multicolumn{13}{l}{\textit{LALM}} \\
StepAudio 3 & 100.0 & 100.0 & 0.0 & 100.0 & 100.0 & 0.0 & 100.0 & 97.1 & 2.9 & 100.0 & 28.5 & 71.5 \\
Audio Flamingo 3 & 99.7 & 98.7 & 1.0 & 99.7 & 4.8 & 94.9 & 100.0 & 94.2 & 5.8 & 100.0 & 23.4 & 76.6 \\
MiMo-Audio & 99.7 & 58.8 & 40.9 & 99.7 & 18.8 & 80.8 & 96.8 & 53.8 & 42.9 & 96.8 & 7.1 & 89.7 \\
\bottomrule
\end{tabular*}
\par\smallskip
\begin{minipage}{\textwidth}\footnotesize
Valid $n$ per operation: omni-modal 625; LVLM 312/313 and LALM 313/312 for task-/evidence-carrying history. Exceptions: Qwen3.8-Omni-Flash Reconfirm 623; Qwen3.8-Omni-Flash Reground 624.
\end{minipage}
\end{table}

History harm counts ST-correct/MT-incorrect pairs; rescue counts the reverse, giving $\Delta_{\mathrm{conv}}=100(\mathrm{harm}-\mathrm{rescue})/n$. Intervention rescue/harm instead compares two multi-turn conditions.

Counts are pooled over both modalities for omni-modal models, visual tasks for LVLMs, and audio tasks for LALMs. The main table separates interfaces; analyses that average them use $(s_{\rm OpenQA}+s_{\rm MCQ})/2$, with each rate computed on its own valid denominator. Figure~\ref{fig:profiles}a pools horizons within models and weights the 13 models equally. Its operation-pair contrasts use the common intersection of valid ST/MT responses for both members, which cannot be recovered from marginal accuracies alone.

\FloatBarrier
\section{Scoring Rules and Automated Sensitivity Audits}
\label{app:scoring}

\paragraph{Frozen rule and content scores.}
The initial scorer applies frozen answer and interface rules, including approved normalization/aliases where applicable. Rule-rejected answers receive blind GPT-5.6 Sol review using the complete question, gold answer, options, and requested answer granularity; model identity, condition, and aggregate performance are hidden. Content scores incorporate confirmed semantic correctness, while uncertain judgments receive no confirmed-correct credit. Rule scores, content scores, format compliance, uncertainty, and review reasons are retained separately. The gold answer and aliases are not modified in response to final test errors. Table~\ref{tab:app-rule-content} reports $\Delta_{\mathrm{conv}}$ under both scoring levels. The production rubric accepts semantic equivalence at the granularity requested by the question; unrequested modifiers, such as finer color shades, are not mandatory solely because they appear in the reference. A text judge assesses answer equivalence, not the validity of the original audiovisual evidence.

\begin{table}[!htbp]
\centering
\small
\setlength{\tabcolsep}{3pt}
\renewcommand{\arraystretch}{1.05}
\caption{\textbf{Frozen rule and reviewed content scoring.} Rule and content scores use the same valid Test single-turn/multi-turn response pairs. Conversational degradation $\Delta_{\mathrm{conv}}$ is in pp; a rule score is not necessarily an exact-string score.}
\label{tab:app-rule-content}
\begin{tabular*}{\textwidth}{@{\extracolsep{\fill}}lrrrrrr@{}}
\toprule
Model & \multicolumn{3}{c}{OpenQA} & \multicolumn{3}{c}{MCQ} \\
\cmidrule(lr){2-4}\cmidrule(l){5-7}
 & $n$ & Rule $\Delta_{\mathrm{conv}}$ & Content $\Delta_{\mathrm{conv}}$ & $n$ & Rule $\Delta_{\mathrm{conv}}$ & Content $\Delta_{\mathrm{conv}}$ \\
\midrule
Qwen3.8-Omni-Flash & 4994 & 9.5 & 7.6 & 4997 & 7.7 & 7.0 \\
Qwen3-Omni & 5000 & 13.9 & 17.3 & 5000 & 21.9 & 21.8 \\
Doubao-Seed-2.0-Lite & 5000 & 19.2 & 20.1 & 5000 & 10.2 & 21.7 \\
MiniCPM-o 4.5 & 5000 & -0.9 & 32.5 & 5000 & 42.5 & 39.5 \\
MiMo-V2.5 & 5000 & 3.2 & 28.0 & 5000 & 29.3 & 33.0 \\
Nemotron-3-Nano-Omni & 5000 & 22.8 & 28.0 & 5000 & 25.0 & 29.0 \\
Qwen3.7-Plus & 2500 & -2.1 & 6.7 & 2500 & 42.0 & 9.6 \\
Qwen3-VL & 2500 & 16.2 & 18.8 & 2500 & 0.0 & 27.8 \\
MiniCPM-V 4.5 & 2500 & -8.1 & 15.6 & 2500 & 44.2 & 26.4 \\
InternVL3.5 & 2499 & 16.5 & 41.1 & 2500 & 30.9 & 30.7 \\
StepAudio 3 & 2498 & 17.1 & 16.2 & 2496 & 17.1 & 16.9 \\
Audio Flamingo 3 & 2500 & 20.3 & 32.4 & 2500 & 41.6 & 39.3 \\
MiMo-Audio & 2500 & 29.0 & 45.2 & 2500 & 55.9 & 55.9 \\
\bottomrule
\end{tabular*}
\end{table}

\begin{table}[!htbp]
\centering
\small
\setlength{\tabcolsep}{4pt}
\renewcommand{\arraystretch}{1.05}
\caption{\textbf{Cross-judge/rubric OpenQA sensitivity audit.} Accuracy estimates (\%) use stratified sampling weights. $n$ counts sampled logical outputs: 480 in total, from 367 distinct judge inputs. Both judge and rubric differ; these estimates do not replace full-Test scores.}
\label{tab:app-alt-scoring}
\begin{tabular*}{\textwidth}{@{\extracolsep{\fill}}llrrrr@{}}
\toprule
Model & Condition & $n$ & Original & Alternative & Disagreements \\
\midrule
Qwen3-Omni & ST & 80 & 94.60 & 82.79 & 17 \\
Qwen3-Omni & MT & 80 & 77.26 & 71.96 & 9 \\
MiniCPM-o 4.5 & ST & 80 & 93.08 & 83.50 & 2 \\
MiniCPM-o 4.5 & MT & 80 & 60.62 & 54.28 & 3 \\
Qwen3.8-Omni-Flash & ST & 80 & 94.76 & 81.86 & 14 \\
Qwen3.8-Omni-Flash & MT & 80 & 87.14 & 76.25 & 11 \\
\bottomrule
\end{tabular*}
\end{table}

\paragraph{Cross-judge and rubric sensitivity.}
The alternative scoring audit retains a positive conversational degradation for all three models, with smaller estimated gaps. Because it changes both judge and rubric, it measures sensitivity to their joint choice rather than isolating a judge effect. We sample 80 logical Test OpenQA outputs in each of six model/condition groups, stratified by rule-correct, review-recovered, still-incorrect, and uncertain outcomes. Deduplication yields 367 distinct Doubao-Seed-2.0-Lite judge inputs for 480 logical outputs. The judge does not see the evaluated model, condition, or original score. Table~\ref{tab:app-alt-scoring} reports sampling-weighted accuracy estimates and unweighted disagreements. The original/alternative weighted $\Delta_{\mathrm{conv}}$ estimates are 17.34/10.82 pp for Qwen3-Omni, 32.46/29.23 pp for MiniCPM-o 4.5, and 7.62/5.61 pp for Qwen3.8-Omni-Flash.

The alternative prompt simplifies the production rules for semantic equivalence and requested answer granularity. All 56 logical disagreements (50 distinct inputs) were originally correct and rejected by the alternative judge. Some involve synonyms, basic color categories, or more specific answers; text-only disagreements do not establish media-grounded error. For MiniCPM-o single-turn, two sampled disagreements account for the 9.58 pp estimated accuracy reduction because each carries 4.79 pp of sampling weight. The single-group design standard error is 6.56 pp, excluding rubric uncertainty. These estimates are not adjudicated scoring-error rates. Shared judge inputs are not independent replications; no cross-condition confidence interval is inferred.

\paragraph{Media-grounded checks.}
An additional 32-pair/64-task sample contains 31 distinct necessary-evidence questions. Qwen3.8-Omni-Flash answers these from the complete media without seeing the gold; 30/31 agree with it. The remaining question asks for the first two spoken words: the gold is ``what is,'' Qwen3.8-Omni-Flash responds ``Most important,'' and an additional blind Doubao-Seed-2.0-Lite media check responds ``What is.'' We retain this disagreement. The paired single-turn input/gold equality and the 64 tasks' option uniqueness and valid answer indices pass programmatic checks. This sample checks necessary evidence, not every competing endpoint, and does not estimate the gold-answer error rate of the full dataset. No human review is included in this audit; Doubao-Seed-2.0-Lite's construction role and Qwen3.8-Omni-Flash's evaluation role also mean that these model checks are not independent ground truth.

\section{Additional Diagnostic Protocols and Results}
\label{app:diagnostics}

\subsection{Task Recall, Restatement, and Historical Answer Text}

The recall probe elicits the historical question; restatement supplies it at the final turn. Table~\ref{tab:app-probes} reports Reground-specific recall denominators. Qwen3-Omni and Qwen3.8-Omni-Flash show high recall despite lower task accuracy, whereas MiniCPM-o 4.5 struggles with recall itself. These separate requests measure elicited recall, not whether recall failed in each original error.

\begin{table}[!htbp]
\centering
\small
\setlength{\tabcolsep}{4pt}
\renewcommand{\arraystretch}{1.05}
\caption{\textbf{Task-recall and historical source-fact probes.} Entries are correct/valid responses on Test-derived probes. The recall column is restricted to Reground. Source-fact probes ask a distinct historical-evidence question; their 712-task cohort is not the recall or full evaluation cohort.}
\label{tab:app-probes}
\begin{tabular*}{\textwidth}{@{\extracolsep{\fill}}lllrr@{}}
\toprule
Model & Interface & Reground recall & Source-fact ST & Source-fact MT \\
\midrule
Qwen3-Omni & OpenQA & 1210/1250 & 656/712 & 532/712 \\
Qwen3-Omni & MCQ & 1226/1250 & 708/712 & 640/712 \\
MiniCPM-o 4.5 & OpenQA & 756/1250 & 644/712 & 476/712 \\
MiniCPM-o 4.5 & MCQ & 940/1250 & 692/712 & 582/712 \\
Qwen3.8-Omni-Flash & OpenQA & 1246/1248 & 652/712 & 610/712 \\
Qwen3.8-Omni-Flash & MCQ & 1239/1250 & 696/712 & 676/712 \\
\bottomrule
\end{tabular*}
\end{table}

Table~\ref{tab:app-interventions} compares question restatement, historical-answer neutralization, and current-media removal on common valid responses. MCQ restatement also supplies the original options, motivating the component study below. Neutralization is not a consistent repair: Qwen3-Omni Reground accuracy falls by 2.64 pp on OpenQA and 13.20 pp on MCQ. Together, the interventions motivate examining task application alongside outdated-answer persistence.

\begin{table}[!htbp]
\centering
\small
\setlength{\tabcolsep}{3.2pt}
\renewcommand{\arraystretch}{1.05}
\caption{\textbf{Component interventions on common valid test responses.} Native and changed accuracies are percentages; $\Delta$ is Changed minus Native. Rescue/harm counts compare the changed response with the native multi-turn response, not single-turn with multi-turn.}
\label{tab:app-interventions}
\begin{tabular*}{\textwidth}{@{\extracolsep{\fill}}llrrrrr@{}}
\toprule
\multicolumn{7}{@{}l}{\textbf{Reground: explicit question restatement}} \\
Model & Interface & $n$ & Native & Changed & $\Delta$ (pp) & Rescue/harm \\
Qwen3-Omni & OpenQA & 1250 & 81.76 & 95.20 & +13.44 & 180/12 \\
Qwen3-Omni & MCQ & 1250 & 57.76 & 98.80 & +41.04 & 514/1 \\
MiniCPM-o 4.5 & OpenQA & 1250 & 58.32 & 92.64 & +34.32 & 441/12 \\
MiniCPM-o 4.5 & MCQ & 1250 & 28.00 & 97.04 & +69.04 & 867/4 \\
Qwen3.8-Omni-Flash & OpenQA & 1249 & 84.07 & 96.40 & +12.33 & 159/5 \\
Qwen3.8-Omni-Flash & MCQ & 1249 & 91.91 & 98.72 & +6.81 & 86/1 \\
\midrule
\multicolumn{7}{@{}l}{\textbf{Reground: neutralize historical answer}} \\
Model & Interface & $n$ & Native & Changed & $\Delta$ (pp) & Rescue/harm \\
Qwen3-Omni & OpenQA & 1250 & 81.76 & 79.12 & -2.64 & 42/75 \\
Qwen3-Omni & MCQ & 1250 & 57.76 & 44.56 & -13.20 & 21/186 \\
MiniCPM-o 4.5 & OpenQA & 1250 & 58.32 & 58.80 & +0.48 & 66/60 \\
MiniCPM-o 4.5 & MCQ & 1250 & 28.00 & 27.84 & -0.16 & 60/62 \\
Qwen3.8-Omni-Flash & OpenQA & 1245 & 84.10 & 71.00 & -13.09 & 46/209 \\
Qwen3.8-Omni-Flash & MCQ & 1248 & 91.91 & 89.34 & -2.56 & 21/53 \\
\midrule
\multicolumn{7}{@{}l}{\textbf{Rebind: remove current media}} \\
Model & Interface & $n$ & Native & Changed & $\Delta$ (pp) & Rescue/harm \\
Qwen3-Omni & OpenQA & 1250 & 52.72 & 92.24 & +39.52 & 502/8 \\
Qwen3-Omni & MCQ & 1250 & 59.60 & 98.24 & +38.64 & 484/1 \\
MiniCPM-o 4.5 & OpenQA & 1250 & 34.96 & 87.84 & +52.88 & 676/15 \\
MiniCPM-o 4.5 & MCQ & 1250 & 31.36 & 95.20 & +63.84 & 804/6 \\
Qwen3.8-Omni-Flash & OpenQA & 1249 & 77.58 & 93.76 & +16.17 & 219/17 \\
Qwen3.8-Omni-Flash & MCQ & 1250 & 79.36 & 99.12 & +19.76 & 251/4 \\
\bottomrule
\end{tabular*}
\end{table}

\subsection{Separating Question and Option Restatement}

We sample 128 Test Reconfirm/Reground pairs (256 tasks per model; seed 2801), with 32 pairs per modality/horizon cell. All arms share media, native prefixes, SDPA, decoding, and both output-format instructions. Components omitted from the final turn remain in history. The reference arm was rerun to match formatting; the other three arms were reused unchanged. Thus $Q_0O_0$ in Table~\ref{tab:app-question-options} is a format-matched diagnostic baseline, not the main-table native prompt.

\begin{table}[!htbp]
\centering
\small
\setlength{\tabcolsep}{4pt}
\renewcommand{\arraystretch}{1.05}
\caption{\textbf{Format-matched question/option restatement.} Entries count correct MCQ answers among $n$ Test tasks per row; Overall pools both operations. $Q_1$ repeats the historical question at the final turn; $O_1$ repeats its original options. All four arms have identical output-format instructions.}
\label{tab:app-question-options}
\begin{tabular*}{\textwidth}{@{\extracolsep{\fill}}llrrrrr@{}}
\toprule
Model & Operation & $n$ & $Q_0O_0$ & $Q_1O_0$ & $Q_0O_1$ & $Q_1O_1$ \\
\midrule
Qwen3-Omni & Reconfirm & 128 & 125 & 121 & 127 & 128 \\
Qwen3-Omni & Reground & 128 & 74 & 93 & 127 & 126 \\
Qwen3-Omni & Overall & 256 & 199 & 214 & 254 & 254 \\
MiniCPM-o 4.5 & Reconfirm & 128 & 110 & 116 & 126 & 125 \\
MiniCPM-o 4.5 & Reground & 128 & 38 & 54 & 118 & 123 \\
MiniCPM-o 4.5 & Overall & 256 & 148 & 170 & 244 & 248 \\
\bottomrule
\end{tabular*}
\end{table}

\noindent\begin{minipage}{\textwidth}
Let $a_{qo}$ denote accuracy, with $q,o\in\{0,1\}$ indicating question and option restatement. Their respective effects are $[(a_{10}-a_{00})+(a_{11}-a_{01})]/2$ and $[(a_{01}-a_{00})+(a_{11}-a_{10})]/2$. On Reground, they are 7.03/33.59 pp for Qwen3-Omni and 8.20/58.20 pp for MiniCPM-o 4.5. The descriptive interactions, $a_{11}-a_{10}-a_{01}+a_{00}$, are $-15.62$ and $-8.59$ pp. Repeating options may assist option inheritance, letter mapping, or task interpretation; the study does not separate these mechanisms.
\end{minipage}

\subsection{Competing Media and Source-Fact Access}

Removing current media reduces both competition and input load. The swap probe retains a competitor but replaces it with a clip supporting a different answer, holding the historical target fixed. Here, the replacement answer is the answer supported by the new current media, not the historical target. Table~\ref{tab:app-value-following} counts strict matches to the replacement answer; its MCQ denominator includes only answers expressible by the original options, unlike the larger swap-accuracy cohort. Qwen3-Omni OpenQA target accuracy changes from 539/1,024 to 529/1,024, while replacement-answer matches rise from 1 to 374. Accuracy can therefore conceal changes in wrong-answer content. The swap changes multiple perceptual attributes, not one isolated value.

\begin{table}[!htbp]
\centering
\small
\setlength{\tabcolsep}{4pt}
\renewcommand{\arraystretch}{1.05}
\caption{\textbf{Following a newly introduced current-media answer in Rebind.} Counts use strict matches to the replacement answer, both before and after the swap. Adopted/abandoned counts transitions into/out of such a match. MCQ uses only cases where the new value is expressible by the original options; its denominator therefore differs from swap-accuracy evaluation.}
\label{tab:app-value-following}
\begin{tabular*}{\textwidth}{@{\extracolsep{\fill}}llrrrr@{}}
\toprule
Model & Interface & $n$ & Before swap & After swap & Adopted/abandoned \\
\midrule
Qwen3-Omni & OpenQA & 1024 & 1 & 374 & 373/0 \\
Qwen3-Omni & MCQ & 980 & 1 & 400 & 399/0 \\
MiniCPM-o 4.5 & OpenQA & 1024 & 2 & 97 & 96/1 \\
MiniCPM-o 4.5 & MCQ & 980 & 0 & 34 & 34/0 \\
Qwen3.8-Omni-Flash & OpenQA & 1022 & 2 & 136 & 134/0 \\
Qwen3.8-Omni-Flash & MCQ & 980 & 1 & 212 & 211/0 \\
\bottomrule
\end{tabular*}
\end{table}

\noindent\begin{minipage}{\textwidth}
The answerability study samples 128 Rebind OpenQA tasks (seed 2804), with 32 per modality/horizon cell. The target, original competitor, and replacement are each queried with the same complete question, using exact cached endpoint requests; three tasks share competitor endpoints from development caches. Table~\ref{tab:app-answerability} reports the full sample and the post-hoc subset with all three endpoints correct. The latter is a sensitivity analysis, not a replacement estimate.
\end{minipage}

\begin{table}[!htbp]
\centering
\small
\setlength{\tabcolsep}{4pt}
\renewcommand{\arraystretch}{1.05}
\caption{\textbf{Competitor-following sensitivity to media answerability.} The first three columns count correct single-turn (ST) answers for the target, old competitor, and new competitor, queried separately with the same question. The final columns count post-swap matches to the new competitor; before-swap counts are zero in every row. The all-three-correct subset accompanies the full 128-task estimate.}
\label{tab:app-answerability}
\begin{tabular*}{\textwidth}{@{\extracolsep{\fill}}lrrrrr@{}}
\toprule
Model & Target ST & Old ST & New ST & All selected & All three correct \\
\midrule
Qwen3-Omni & 120/128 & 120/128 & 123/128 & 40/128 & 37/111 \\
MiniCPM-o 4.5 & 113/128 & 122/128 & 121/128 & 6/128 & 4/107 \\
Qwen3.8-Omni-Flash & 119/128 & 119/128 & 117/128 & 13/128 & 12/102 \\
\bottomrule
\end{tabular*}
\end{table}

Historical source-fact probes ask a different question about the referenced media (Table~\ref{tab:app-probes}; 712 valid probes per model/interface). Excluding literal gold strings in historical assistant text retains 624, 590, and 620 tasks for Qwen3-Omni, MiniCPM-o 4.5, and Qwen3.8-Omni-Flash. Other textual cues may remain; these probes complement task accuracy rather than decompose individual errors.

\subsection{Historical Slots and Explicit Reference Wording}

The slot study uses 932 Test-derived tasks with eligible, sometimes reused donors. Near removes a historical media/assistant slot; Far-Neutral inserts one; Far-Competing replaces its media and reply with a competing donor. For Rebind, 400 targets at positions two through four are moved to the first position, changing their ordinal cue. Table~\ref{tab:app-history-slots} uses common valid denominators within each operation/interface. Content, load, replies, and sometimes position co-vary, so this is not a pure distance experiment.

\begin{table}[!htbp]
\centering
\small
\setlength{\tabcolsep}{3pt}
\renewcommand{\arraystretch}{1.05}
\caption{\textbf{History-slot interventions on a donor-eligible test-derived cohort.} Values are accuracy (\%). Far-N inserts a neutral slot and Far-C replaces it with a competing slot. Near removes this slot. These conditions vary content/load as well as target distance.}
\label{tab:app-history-slots}
\begin{tabular*}{\textwidth}{@{\extracolsep{\fill}}lllrrrrr@{}}
\toprule
Model & Interface & Operation & $n$ & ST & Near & Far-N & Far-C \\
\midrule
Qwen3-Omni & OpenQA & Reground & 400 & 97.0 & 81.8 & 71.8 & 75.5 \\
Qwen3-Omni & MCQ & Reground & 400 & 99.8 & 43.8 & 43.2 & 47.2 \\
Qwen3-Omni & OpenQA & Rebind & 531 & 94.9 & 79.7 & 44.4 & 31.1 \\
Qwen3-Omni & MCQ & Rebind & 531 & 99.8 & 92.1 & 74.2 & 63.5 \\
Qwen3.8-Omni-Flash & OpenQA & Reground & 399 & 98.5 & 82.5 & 75.7 & 78.2 \\
Qwen3.8-Omni-Flash & MCQ & Reground & 400 & 99.5 & 92.2 & 89.5 & 91.0 \\
Qwen3.8-Omni-Flash & OpenQA & Rebind & 531 & 94.2 & 71.9 & 71.9 & 66.1 \\
Qwen3.8-Omni-Flash & MCQ & Rebind & 530 & 99.1 & 72.6 & 72.3 & 71.7 \\
\bottomrule
\end{tabular*}
\end{table}

\noindent\begin{minipage}{\textwidth}
The wording study samples 64 pairs/128 OpenQA tasks (seed 2805; 32 tasks per operation). Two predefined rewrites refer to the message or explicitly to its attached media, preserving target ordinals. Task-carrying requests still inherit the first question and use current media; they do not restate the factual question. Table~\ref{tab:app-reference-rewrites} shows heterogeneous changes and lower overall accuracy. Original responses were cached from the same hosted endpoint and native prefix; endpoint drift and residual ambiguity prevent a causal interpretation of these wording effects.
\end{minipage}

\begin{table}[!htbp]
\centering
\small
\setlength{\tabcolsep}{4pt}
\renewcommand{\arraystretch}{1.05}
\caption{\textbf{Two fixed reference rewrites for Qwen3.8-Omni-Flash.} Entries are correct OpenQA answers; R/H denotes method rescue/harm relative to cached original responses. The 128 tasks comprise 64 matched pairs.}
\label{tab:app-reference-rewrites}
\begin{tabular*}{\textwidth}{@{\extracolsep{\fill}}lrrrrrr@{}}
\toprule
Operation & $n$ & Original & Message & R/H & Explicit media & R/H \\
\midrule
Reconfirm & 32 & 31 & 26 & 0/5 & 25 & 0/6 \\
Reground & 32 & 26 & 24 & 1/3 & 24 & 3/5 \\
Retrieve & 32 & 30 & 30 & 0/0 & 30 & 0/0 \\
Rebind & 32 & 25 & 26 & 4/3 & 28 & 4/1 \\
Overall & 128 & 112 & 106 & 5/11 & 107 & 7/12 \\
\bottomrule
\end{tabular*}
\end{table}

\section{Adaptation Protocol and Additional Results}
\label{app:adaptation}

\paragraph{Training data and candidates.}
Qwen3-Omni is adapted on 1,018 Short OpenQA views in 509 pairs, derived from 750 unique Dev-Train task IDs. There are 375 answer-preserving pairs and 134 target-switching pairs. Up to three verified candidate values are used per pair; duplicate detection strips and collapses whitespace and applies Unicode case folding while retaining the frozen display value for tokenization. Candidate scores average answer-token log probabilities, excluding prompt padding and EOS. Among answer-preserving pairs, 284 have at least two candidates for the JS term; the other 91 receive answer-token cross-entropy only. Historical text literally contains the relevant gold in 74/134 first-target views of the target-switching set (59/73 audio and 15/61 visual), so these views do not universally require media re-access.

For eligible answer-preserving pairs, the consistency-augmented objective is
\begin{equation}
\mathcal{L}_{\mathrm{SFT+Inv}}=\mathcal{L}_{\mathrm{CE}}+
\lambda\,\mathrm{JS}\!\left(p_\theta(\cdot\mid h_i),p_\theta(\cdot\mid h_j)\right),
\label{eq:invariance}
\end{equation}
where the distributions are over the verified candidate set and $\lambda=0.1$. The SFT control sets the auxiliary coefficient to zero while retaining the auxiliary-forward budget. The candidate set is used during training, not supplied to the model at test time.

\paragraph{Matched optimization.}
SFT and SFT+Inv independently start from the same base checkpoint with seed 53126, identical data order, and identical LoRA initialization. LoRA~\citep{hu2021lora} targets text-attention query/key/value/output projections in layers 12--47 with rank 8, scale 16, dropout 0, and no bias adaptation. AdamW~\citep{loshchilov2019decoupled} uses learning rate $10^{-5}$, betas $(0.9,0.999)$, epsilon $10^{-8}$, no weight decay, a constant schedule, no warmup, and gradient clipping at 1.0. Global batch size is 5 for 204 updates, giving 1,020 answer exposures and 3,352 candidate forwards per arm, on two GPUs in one model-parallel rank. Both arms execute the same candidate forwards/backwards; SFT sets the auxiliary coefficient to zero. Cross-entropy is accumulated per answer view and the pair auxiliary term once per left-view exposure, divided by the same global batch size rather than separately normalized by pair count.

\begin{table}[!htbp]
\centering
\small
\setlength{\tabcolsep}{4pt}
\renewcommand{\arraystretch}{1.05}
\caption{\textbf{Qwen3-Omni adaptation on Dev-Val.} Accuracy (\%) on 500 tasks per interface. The additional sensitivity objective was evaluated on Dev-Val only.}
\label{tab:app-training-val}
\begin{tabular*}{\textwidth}{@{\extracolsep{\fill}}lrrrr@{}}
\toprule
Training objective & \multicolumn{2}{c}{OpenQA} & \multicolumn{2}{c}{MCQ} \\
\cmidrule(lr){2-3}\cmidrule(l){4-5}
 & ST & MT & ST & MT \\
\midrule
SFT & 90.8 & 74.8 & 96.8 & 80.8 \\
SFT+Inv & 90.0 & 75.0 & 96.8 & 80.6 \\
SFT+Inv+sensitivity & 90.8 & 74.4 & 96.8 & 81.0 \\
\bottomrule
\end{tabular*}
\end{table}

\begin{table}[!htbp]
\centering
\small
\setlength{\tabcolsep}{3pt}
\renewcommand{\arraystretch}{1.05}
\caption{\textbf{Qwen3-Omni native-history adaptation on Test.} Accuracy (\%); $\Delta_{\mathrm{conv}}$ is in pp. Checkpoints generate their own histories on the same valid tasks. SFT and SFT+Inv share examples, initialization, updates, and auxiliary-forward budget.}
\label{tab:app-training-test}
\begin{tabular*}{\textwidth}{@{\extracolsep{\fill}}lrrrrrrr@{}}
\toprule
Metric / subset & $n$ & \multicolumn{3}{c}{OpenQA} & \multicolumn{3}{c}{MCQ} \\
\cmidrule(lr){3-5}\cmidrule(l){6-8}
 &  & Base & SFT & SFT+Inv & Base & SFT & SFT+Inv \\
\midrule
Overall ST & 5000 & 94.60 & 94.84 & 94.92 & 99.12 & 99.12 & 99.12 \\
Overall MT & 5000 & 77.26 & 80.68 & 80.70 & 77.28 & 79.64 & 79.84 \\
Overall $\Delta_{\mathrm{conv}}$ & 5000 & 17.34 & 14.16 & 14.22 & 21.84 & 19.48 & 19.28 \\
Reconfirm MT & 1250 & 90.08 & 91.68 & 91.92 & 96.40 & 95.76 & 95.68 \\
Reground MT & 1250 & 81.76 & 85.84 & 85.52 & 57.76 & 62.96 & 63.36 \\
Retrieve MT & 1250 & 84.48 & 85.36 & 85.20 & 95.36 & 95.76 & 96.24 \\
Rebind MT & 1250 & 52.72 & 59.84 & 60.16 & 59.60 & 64.08 & 64.08 \\
Short MT & 2500 & 91.64 & 94.12 & 94.36 & 89.68 & 90.72 & 90.76 \\
Long MT & 2500 & 62.88 & 67.24 & 67.04 & 64.88 & 68.56 & 68.92 \\
\bottomrule
\end{tabular*}
\end{table}

\begin{table}[!htbp]
\centering
\small
\setlength{\tabcolsep}{4pt}
\renewcommand{\arraystretch}{1.05}
\caption{\textbf{Qwen3-Omni paired changes on Test.} $n=5,000$ per row; $\Delta$ subtracts the second method's multi-turn accuracy from the first. 95\% percentile intervals use 2,000 derivation-root resamples (1,875 roots); parent media need not be independent.}
\label{tab:app-training-comparisons}
\begin{tabular*}{\textwidth}{@{\extracolsep{\fill}}llrrl@{}}
\toprule
Comparison & Interface & $\Delta$ (pp) & Rescue/harm & Root interval \\
\midrule
SFT vs. Base & OpenQA & +3.42 & 261/90 & $[2.60, 4.23]$ \\
SFT vs. Base & MCQ & +2.36 & 163/45 & $[1.80, 2.90]$ \\
SFT+Inv vs. SFT & OpenQA & +0.02 & 26/25 & $[-0.24, 0.30]$ \\
SFT+Inv vs. SFT & MCQ & +0.20 & 41/31 & $[-0.14, 0.54]$ \\
\bottomrule
\end{tabular*}
\end{table}

\noindent\begin{minipage}{\textwidth}
\paragraph{Selection and interpretation.}
A Dev-Val-only sensitivity-hinge variant ($\lambda=0.1$, margin 0.2) was not evaluated on Test (Table~\ref{tab:app-training-val}). SFT and SFT+Inv generate native test histories without candidate reranking. Tables~\ref{tab:app-training-test} and~\ref{tab:app-training-comparisons} report accuracy, repair/harm counts, and conditional derivation-root intervals. SFT reduces MCQ Reconfirm accuracy by 0.64 pp; JS adds only one correct OpenQA answer and ten MCQ answers over SFT, with intervals spanning zero. The observed improvement comes from SFT, with no clear incremental gain from consistency regularization. Development includes mixed horizons and interfaces, so Long/MCQ are not wholly unseen conditions. The study uses one model and seed, without an unpaired-data control.
\end{minipage}

\paragraph{Source-overlap sensitivity.}
Metadata-only filtering excludes Test tasks sharing any input-media parent with gradient-training inputs, retaining 3,125 tasks per interface (Table~\ref{tab:app-training-source}). SFT gains remain 3.14 pp on OpenQA (148 repairs, 50 harms) and 2.50 pp on MCQ (106 repairs, 28 harms), so improvement is not confined to shared training parents. Filtering changes task composition and does not remove all development exposure. The all-input Dev-disjoint subset contains no Long Retrieve/Rebind tasks, so its gains describe that restricted task mixture.

\begin{table}[!htbp]
\centering\small
\setlength{\tabcolsep}{3pt}
\caption{\textbf{Qwen3-Omni SFT gains under source-overlap exclusions.} Training refers to actual gradient-training inputs. Each subset is fixed without reading model scores; $n$ is common to both interfaces. Gains are SFT minus Base multi-turn accuracy (pp); R/H counts repaired/harmed answers. Source filtering changes the task mixture.}
\label{tab:app-training-source}
\begin{tabular*}{\textwidth}{@{\extracolsep{\fill}}lrrrrr@{}}
\toprule
\multirow{2}{*}{Retained subset} & \multirow{2}{*}{$n$} & \multicolumn{2}{c}{OpenQA} & \multicolumn{2}{c}{MCQ} \\
\cmidrule(lr){3-4}\cmidrule(l){5-6}
 & & Gain & R/H & Gain & R/H \\
\midrule
All Test tasks & 5,000 & +3.42 & 261/90 & +2.36 & 163/45 \\
Target parent disjoint from Dev & 4,082 & +3.58 & 218/72 & +2.13 & 127/40 \\
All input parents disjoint from Dev & 1,928 & +2.65 & 78/27 & +1.66 & 54/22 \\
Target parent disjoint from training & 4,514 & +3.57 & 239/78 & +2.10 & 139/44 \\
All input parents disjoint from training & 3,125 & +3.14 & 148/50 & +2.50 & 106/28 \\
\bottomrule
\end{tabular*}
\end{table}

\section{Inference-Time Assistance: Test and Dev-Val Protocols}
\label{app:inference}

\paragraph{Actions and routing.}
Restatement supplies the inherited question and original MCQ options. Reweighting resolves the explicit target-media reference, scales non-target media and historical assistant-text keys by 0.5 at final-answer positions, and renormalizes attention. All inputs remain available, and historical representations are unchanged. A deterministic parser routes historical-question requests to restatement and complete-question requests to reweighting, with neutral fallback for unresolved references. It uses neither gold answers nor operation labels.

\paragraph{Test action comparison.}
On 5,000 Test tasks per interface with cached Qwen3-Omni base histories, routing rescues/harms 140/13 OpenQA answers (+2.54 pp) and 21/10 MCQ answers (+0.22 pp) relative to restatement (Table~\ref{tab:app-inference-test}). Routing reuses selected outputs without another forward pass. Actions share SDPA, whereas the formal native baseline uses its original runtime; comparisons with that baseline cannot isolate intervention effects or rank assistance against adaptation.

\begin{table}[!htbp]
\centering
\small
\setlength{\tabcolsep}{4pt}
\renewcommand{\arraystretch}{1.05}
\caption{\textbf{Qwen3-Omni inference actions on Test.} Accuracy (\%), with $n=5,000$ Test tasks per interface. All actions share SDPA execution and cached base-model histories. ST is their shared neutral single-turn reference; $\Delta_{\mathrm{conv}}$ is ST$-$MT (pp). This table compares actions; the runtime-matched neutral multi-turn comparison is reported separately on Dev-Val.}
\label{tab:app-inference-test}
\begin{tabular*}{\textwidth}{@{\extracolsep{\fill}}lrrrrrr@{}}
\toprule
Action & \multicolumn{3}{c}{OpenQA} & \multicolumn{3}{c}{MCQ} \\
\cmidrule(lr){2-4}\cmidrule(l){5-7}
 & ST & MT & $\Delta_{\mathrm{conv}}$ & ST & MT & $\Delta_{\mathrm{conv}}$ \\
\midrule
Restatement & 95.00 & 81.48 & 13.52 & 99.12 & 87.80 & 11.32 \\
Reweighting & 95.00 & 79.40 & 15.60 & 99.12 & 77.32 & 21.80 \\
Routed & 95.00 & 84.02 & 10.98 & 99.12 & 88.02 & 11.10 \\
\bottomrule
\end{tabular*}
\end{table}

\paragraph{Validation neutral comparison.}
On 500 Dev-Val tasks per model/interface, neutral execution and actions share cached histories, media, SDPA, and decoding (Table~\ref{tab:app-inference-val}). Reweighting changes OpenQA/MCQ accuracy by +1.8/$-1.2$ pp for Qwen3-Omni and +1.6/+0.6 pp for MiniCPM-o 4.5. Routing trails restatement by 0.4 pp on Qwen3-Omni MCQ. MiniCPM-o 4.5's single-turn OpenQA reference is 446/500, versus 456/500 under the earlier native configuration; this is a runtime difference, not a routing effect. This matched neutral ablation covers Dev-Val, not full Test.

\begin{table}[!htbp]
\centering
\small
\setlength{\tabcolsep}{4pt}
\renewcommand{\arraystretch}{1.05}
\caption{\textbf{Runtime-matched neutral comparison on Dev-Val.} Accuracy (\%), with $n=500$ per row. Neutral and all three actions use identical cached histories, media, SDPA execution, and decoding. The neutral ST condition is shared within each row.}
\label{tab:app-inference-val}
\begin{tabular*}{\textwidth}{@{\extracolsep{\fill}}llrrrrr@{}}
\toprule
Model & Interface & ST & Neutral & Restate & Reweight & Routed \\
\midrule
Qwen3-Omni & OpenQA & 89.2 & 71.4 & 76.0 & 73.2 & 78.2 \\
Qwen3-Omni & MCQ & 96.8 & 78.8 & 86.4 & 77.6 & 86.0 \\
MiniCPM-o 4.5 & OpenQA & 89.2 & 59.8 & 74.2 & 61.4 & 77.8 \\
MiniCPM-o 4.5 & MCQ & 99.6 & 59.6 & 78.2 & 60.2 & 78.2 \\
\bottomrule
\end{tabular*}
\end{table}

\section{Scope and Limitations}
\label{app:scope}

ReTurn makes these requirements measurable through controlled task constructions; it does not exhaust open-ended conversation. Its references are explicit, histories are short relative to lifelong interaction, and source media can recur across partitions. Construction and scoring rely on automated checks, with alternative-judge and media-grounded sensitivity audits rather than human agreement estimates. The diagnostics characterize behavioral responses to controlled interventions; broader generalization remains to be assessed. Broader histories, independently held-out media, and replicated adaptation would test how far the present findings extend.

\end{document}